\documentclass{article} 
\usepackage{iclr2026_conference,times}

\usepackage{amsmath,amsfonts,bm}

\def\eqref#1{equation~\ref{#1}}

\def\1{\bm{1}}

\DeclareMathAlphabet{\mathsfit}{\encodingdefault}{\sfdefault}{m}{sl}
\SetMathAlphabet{\mathsfit}{bold}{\encodingdefault}{\sfdefault}{bx}{n}

\usepackage{hyperref}
\usepackage{url}
\usepackage{graphicx}
\usepackage{subcaption}
\usepackage{enumitem}
\usepackage{xspace}
\usepackage{amsmath}
\usepackage{cleveref}
\usepackage{booktabs}
\usepackage{tabularx}
\usepackage{multirow}
\usepackage{array}
\usepackage{colortbl}
\usepackage[table,dvipsnames]{xcolor}

\usepackage{amssymb}
\usepackage{pifont}
\newcommand{\cmark}{\ding{51}}%
\newcommand{\xmark}{\ding{55}}%

\crefname{section}{\S}{\S\S}
\crefname{figure}{fig.}{figs.}%
\crefname{appendix}{app.}{app.}%

\newcommand{\steplevel}{\texttt{Step-Level}\xspace} 
\newcommand{\resplevel}{\texttt{Response-Level}\xspace} 
\newcommand{\errorid}{\texttt{ErrorID}\xspace}

\usepackage[most]{tcolorbox}
\usepackage{listings}
\usepackage{xcolor}
\usepackage{etoolbox} 
\newtcolorbox{prompt}{
  colback=black!2,    
  colframe=black!15,  
  boxrule=0.4pt,
  sharp corners,
  left=8pt,right=8pt,top=6pt,bottom=6pt,
  before skip=6pt, after skip=6pt,
  breakable,                 
  before upper=\ttfamily\small 
}

\definecolor{RubricNavy}{RGB}{16,36,132}    
\definecolor{RubricBack}{RGB}{235,238,249}  
\definecolor{RubricFrame}{RGB}{16,36,132}   

\newtcolorbox{rubricbox}[1]{%
  enhanced, breakable,
  colback=RubricBack,
  colframe=RubricFrame,
  colbacktitle=RubricNavy, coltitle=white,
  title=\sffamily\bfseries #1,
  boxrule=0.8pt, arc=2.5mm,
  left=8pt,right=8pt,top=8pt,bottom=8pt,
  before skip=8pt, after skip=8pt,
  before upper=\ttfamily\small 
                 \setlength{\parindent}{0pt}%
               \setlength{\parskip}{4pt}%

}

\newtoggle{isiclr}
\togglefalse{isiclr}
\iftoggle{isiclr}{
    \newcommand{\turing}{{[Data Co.]}\xspace}
    \newcommand{\turingredact}{{[Data Co.] (Redacted for Review)}\xspace}
}{
    \newcommand{\turing}{{Turing}\xspace}
    \newcommand{\turingredact}{{Turing}\xspace}
}

\title{Hard2Verify: A Step-Level Verification Benchmark for Open-Ended Frontier Math}

\newlength{\leaderboardwith}
\author{
\parbox{\textwidth}{
{
    \centering
    \normalfont
    \vspace{0.5em} 
    \vspace{0.5em} 
    \textbf{Shrey Pandit}$^\star$, \textbf{Austin Xu}$^\star$, \textbf{Xuan-Phi Nguyen}, \textbf{Yifei Ming}, \textbf{Caiming Xiong}, \textbf{Shafiq Joty} \\[0.5em]
    Salesforce AI Research \\
    {\small $^\star$Equal Contribution, \texttt{\{shrey.pandit, austin.xu\}@salesforce.com}\\[1em]}
}
{ 
    \small
    \normalfont
    \makebox[\leaderboardwith][l]{~~~~~~~~~~Data:}\hspace{0em}\url{https://huggingface.co/datasets/Salesforce/Hard2Verify}\\
    \makebox[\leaderboardwith][l]{~~~~~~~~~~Code:}\hspace{0em}\url{https://github.com/SalesforceAIResearch/Hard2Verify}
}}}

\newcommand{\hardtoverify}{\textit{Hard2Verify}\xspace}

\newtoggle{comments}
\toggletrue{comments}

\iftoggle{comments}{
    \newcommand{\shrey}[1]{\textcolor{blue}{(Shrey: #1)}}
    \newcommand{\austin}[1]{\textcolor{orange}{(Austin: #1)}}
    \newcommand{\jason}[1]{\textcolor{violet}{(Peifeng: #1)}}
    \newcommand{\shafiq}[1]{\textcolor{cyan}{(shafiq: #1)}}
    \newcommand{\nxphi}[1]{\textcolor{red}{(Phi: #1)}}
}{
    \newcommand{\shrey}[1]{}
    \newcommand{\austin}[1]{}
    \newcommand{\jason}[1]{}
    \newcommand{\shafiq}[1]{}
    \newcommand{\nxphi}[1]{}
}

\iclrfinalcopy 
\begin{document}

\maketitle

\begin{abstract}
Large language model (LLM)-based reasoning systems have recently achieved gold medal-level performance in the IMO 2025 competition, writing mathematical proofs where, to receive full credit, each step must be not only correct but also sufficiently supported. To train LLM-based reasoners in such challenging, open-ended settings, strong verifiers capable of catching step-level mistakes are necessary prerequisites. We introduce \hardtoverify, a human-annotated, step-level verification benchmark produced with over 500 hours of human labor. Hard2Verify is designed to rigorously assess step-level verifiers at the frontier: Verifiers must provide step-level annotations or identify the first error in responses generated by frontier LLMs for very recent, challenging, and open-ended math questions. We evaluate 29 generative critics and process reward models, demonstrating that, beyond a few standouts, open-source verifiers lag closed source models. We subsequently analyze what drives poor performance in step-level verification, the impacts of scaling verifier compute, as well as fundamental questions such as self-verification and verification-generation dynamics. 
\end{abstract}
\begin{figure}[h!]
    \centering
    \includegraphics[width=\linewidth]{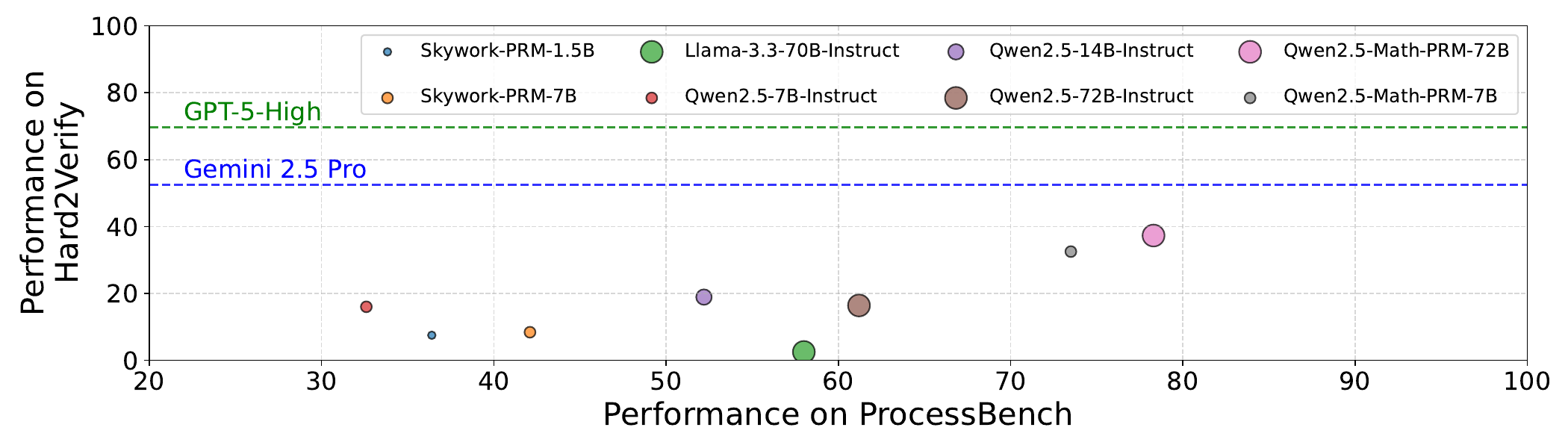}
    \caption{Comparison of models evaluated on both ProcessBench \citep{zheng2024processbench} and our Hard2Verify benchmark. Past benchmarks do not sufficiently evaluate in the frontier-level math settings that Hard2Verify does; On the same error identification task, Qwen2.5-Math-PRM-72B performance drops from ProcessBench state-of-the-art at 78.3 to 37.3 on Hard2Verify.
    }
    \label{fig:teaser}
\end{figure}

\section{Introduction}
Mathematical reasoning serves as a gold-standard evaluation setting for benchmarking reasoning progress in large language models (LLMs). Over the past half-decade, benchmarks have been introduced to assess LLMs at the grade-school~\citep{cobbe2021gsm8k}, high-school~\citep{hendrycks2021measuring}, university~\citep{zhang2023evaluating}, and competition math level~\citep{aime,he2024olympiadbench,gao2024omni}. However, the progress of mathematical reasoning ability of LLMs has outpaced benchmark creation, with every subsequent release of a frontier LLM saturating new benchmarks, most recently with GPT-5 Pro achieving 96.5\%+ on AIME 2024. As a result, recent efforts~\citep{glazer2024frontiermath,phan2025humanity} have written novel, unseen mathematical questions to test LLMs.

While the training approaches of closed frontier models remain a secret, open-source progress in mathematical reasoning has been driven by scaling \textit{reinforcement learning from verifiable rewards} (RLVR)~\citep{lambert2024tulu}, with the breakthrough of DeepSeek-R1~\citep{guo2025deepseek} leading to an explosion of interest. This paradigm requires training data with solutions that are easily \textit{verifiable}, i.e., have solutions that can be easily checked against a known ground-truth by string matching or symbolic checkers. Math benchmarks, for the most part, also adopt the verifiable setup, where a model response is considered correct if its final answer matches the established ground-truth. Answer correctness, while a necessary condition for overall solution correctness, is not sufficient: It is now established that LLMs can produce incorrect intermediate reasoning but conclude with correct final answers~\citep{lightman2023let,zheng2024processbench,setlur2025rewarding}.

The next frontier for LLMs is solving problems that are \textit{hard to verify}. A grand example of such a problem is proving the \emph{Riemann hypothesis}, where the expected solution is not a short phrase, but a multi-step proof. To verify correctness, each step must be rigorously checked. Hints of open-ended problem solving abilities already exist: advanced reasoning systems~\citep{openai_imo,gemini_deep_think_imo,huang2025gemini} have achieved gold-level performance in the 2025 \href{https://www.imo-official.org/}{IMO}. Here, LLM outputs were judged at the step-level by human experts who determined if steps are both correct and sufficiently supported, with supporting lemmas and claims all appropriately stated and applied.

Training reasoning LLMs capable of open-ended problem solving requires scalable \textit{automatic evaluation}: Not every LLM rollout during RLVR training can be audited by human experts. Rather, evaluation in open-ended settings requires \textit{step-level verifiers}, typically process reward models (PRMs) or generative critic models. Such verifiers have already been used to provide dense process rewards \citep{lightman2023let,grpo_shao2024deepseekmath,zha2025rl}. Furthermore, step-level verifiers are also used in many test-time scaling methods, selecting the most promising candidate from multiple solutions or steps \citep{scaling_testtime_optimallysnell2024scaling,scaling_flaws_of_verifier_guided_yu2025scaling,lifshitz2025multi,zhou2025evaluating}. However, are these step-level verifiers sufficient for pushing the frontier of mathematical reasoning?

This work introduces \hardtoverify, which gauges the ability of step-level verifiers to push the frontier. \hardtoverify benchmarks verifiers in assessing \textit{frontier} LLM responses to difficult, recent, and open-ended math problems. We curate challenging problems from recent international mathematics competitions like IMO and \href{http://maa.org/putnam/}{Putnam}, which are used to sample responses from three top-tier LLMs, GPT-5 (high)~\citep{gpt5}, Gemini 2.5 Pro~\citep{gemini_25}, and Claude Sonnet 4 (thinking)~\citep{claude4}. Finally, we employ PhD-level math experts to annotate each model-generated step. The resulting benchmark is the culmination of over 500 hours of human effort, passing three rounds of independent agreement checks. This meticulous process yields 1860 rigorously graded steps across 200 unique model responses.

\begin{table}[t!]
\caption{Comparison between Hard2Verify and existing step-level math benchmarks.}
\label{tab:dataset_comp}
\resizebox{\textwidth}{!}{%
\begin{tabular}{lcccccc}
\toprule
 & \begin{tabular}[c]{@{}c@{}}Question \\ Difficulty\end{tabular} & \begin{tabular}[c]{@{}c@{}}Open-Ended \\ Responses?\end{tabular} & \begin{tabular}[c]{@{}c@{}}Natural \\ Responses?\end{tabular} & \begin{tabular}[c]{@{}c@{}}Generator \\ Strength\end{tabular} & Annotator & \begin{tabular}[c]{@{}c@{}}Step-Level \\ Labels?\end{tabular}\\ 
 \midrule
MR-GSM8K~\citep{zeng2023mr} & Easy & \xmark & \cmark & Weak & Human & \cmark \\
MR-MATH~\citep{xia2025evaluating} & Easy & \xmark & \cmark & Weak & Human & \cmark \\
MR-Ben~\citep{zeng2023mr} & Easy & \xmark & \cmark & Weak & Human & \cmark\\
ProcessBench~\citep{zheng2024processbench} & Easy-Hard & 10.3\% & \cmark & Weak-Medium & Human & \xmark \\
PRMBench~\citep{song2025prmbench} & Easy & \xmark & \xmark & Weak & Synth. + Human Check & \cmark \\ 
\midrule
Hard2Verify (Ours) & Hard & 78.5\% & \cmark & Strong & Human & \cmark \\ 
\bottomrule
\end{tabular}%
}
\end{table}

Beyond operating at the frontier, \hardtoverify distinguishes itself from existing benchmarks for step-level annotation (\Cref{tab:dataset_comp}). First, we emphasize collecting open-ended questions, with 78.5\% of our samples being open-ended. This way, verifiers cannot ``cheat'' if they have seen the question or ground-truth answer during training; rather verifiers must substantively assess step correctness. Second, step correctness is judged not only on correctness, but also based on whether all invoked results, such as supporting lemmas or claims, are correctly stated and applied; saying ``$X$ follows from $Y$'' receives no credit if $Y$ is not sufficiently justified or properly invoked. Third, \hardtoverify focuses on benchmarking verifiers in naturally occurring application settings: Verifiers must assess \textit{model-written} responses, which often differ dramatically from human-written reference answers.

We benchmark 29 models spanning proprietary models to open-weight models to PRMs. Compared to past work, \hardtoverify represents a step up in difficulty, as shown in~\Cref{fig:teaser}; Models capable of scoring 60\%+ on ProcessBench~\citep{zheng2024processbench} are unable to crack 20\% on \hardtoverify. Our analysis reveals that this degraded performance is because weaker verifiers cannot identify mistakes, marking nearly \textit{every} step as correct. We additionally analyze several fundamental questions in step-level verification: How should one to scale verifier compute? What are the impacts of self-verification? How much easier is generation than verification for frontier models?

\section{Background and Related Work}
\textbf{LLM-based verification.} 
To meet demands for scalable evaluation, LLM-based evaluators have been proposed, originally focusing on chat settings~\citep{mtbench_zheng2023judging}. However, as LLMs are deployed in challenging reasoning settings~\citep{ke2025survey}, recent evaluations have shown the need for more capable evaluators in reasoning domains~\citep{frick2024evaluate,tan2024judgebench,zhou2025evaluating}. To get denser evaluation signal, focus quickly shifted to PRMs~\citep{lightman2023let} and synthetic ways to curate step-level training data~\citep{wang2023math,luo2024improve}. However, when used as dense reward signals for policy optimization, recent work has shown only limited improvement over outcome-level counterparts~\citep{grpo_shao2024deepseekmath}, which results from shortcomings process reward formulations. PRMs only measure if a step \textit{could} lead to a correct, likely short-form final answer, not whether the step is correct in any absolute sense. 
As a result, recent focused has shifted towards \textit{generative verifiers}~\citep{mahan2024generative,zhang2025generative,liu2025inference}, using the natural language generation abilities of LLMs to perform verification. This allows for more precise description of evaluation criteria and the ability to increase inference-time compute during the verification process.

\textbf{Benchmarking step-level verifiers in math settings.} \Cref{tab:dataset_comp} contrasts \hardtoverify with related benchmarks. MR-GSM8K~\citep{zeng2023mr} annotate model responses to GSM8K~\citep{cobbe2021gsm8k} questions on a per-step basis to evaluate generative models as evaluators. MR-MATH~\citep{xia2025evaluating} and MR-Ben~\citep{zeng2024mr} follow similar approachs, increasing question difficulty with slightly harder sources like MATH~\citep{hendrycks2021measuring} and MMLU~\citep{hendrycks2020measuring}.
The two most relevant works to \hardtoverify are ProcessBench~\citep{zheng2024processbench} and PRMBench~\citep{song2025prmbench}. ProcessBench uses a mix of easy (GSM8K and MATH) and hard (OlympiadBench and Omni-MATH) questions, but is comprised largely of samples with single answer outputs\footnote{
    To compute the fraction of open-ended questions in ProcessBench reported in~\Cref{tab:dataset_comp}, we count the number questions from the Omni-MATH split that do not appear in the \href{https://github.com/KbsdJames/omni-math-rule}{Omni-MATH rule-based split.} All other ProcessBench splits contain questions that are not open-ended.
}. Further, ProcessBench only evaluates first error identification ability of verifiers, rather than tasking verifiers to evalute \textit{every} step. PRMBench, on the other hand, obtains step-level annotations by taking fully correct \textit{human-written} and model-generated solutions from PRM800K and injecting errors with an LLM, yielding responses that are \textit{not naturally occurring}: Human-and model-written text may have large differences in style and substance, while injected errors may not represent naturally occurring errors in model generation. Hard2Verify, in contrast, operates at the current frontier, tasking verifiers to evaluate responses from frontier-level LLMs to difficult, largely open-ended questions.
\begin{figure}
    \centering
    \includegraphics[width=\linewidth]{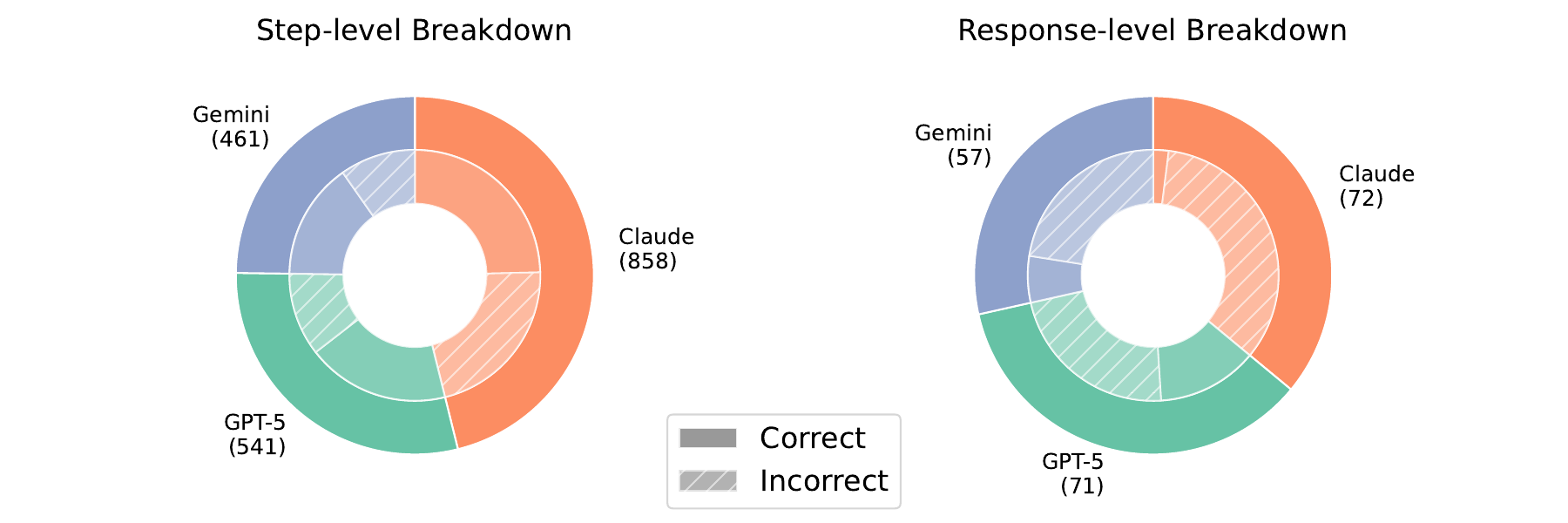}
    \caption{Breakdown of correct vs. incorrect steps (left) and responses (right) by model. We consider a response incorrect if \textit{any} step in the response is labeled incorrect.}
    \label{fig:resp_breakdown}
\end{figure}

\section{The Hard2Verify Benchmark}
\subsection{Design philosophy}
Hard2Verify is designed to test verifiers at the frontier of LLM-based math reasoning. At the question, response, and annotation level, Hard2Verify is curated based on the following philosophy:
\begin{itemize}[leftmargin=*,noitemsep,topsep=1pt]
    \item \textbf{Questions.} To measure progress in step-level verification, we must characterize how verifiers perform on \textit{extremely difficult, open-ended} math questions. Open-ended problems represent the next frontier of mathematical reasoning, one where verifiers become increasingly important in lieu of available ground-truth answers. We focus our data collection on very recent mathematical Olympiads, prioritizing open-ended questions.
    \item \textbf{Model responses.} The responses that verifiers evaluate must be from \textit{highly capable, frontier-level models.} To push the frontier of math reasoning, verifiers must be able to tell when the most powerful models make potentially subtle mistakes. Moreover, such mistakes should be \textit{naturally occurring}, i.e., arise naturally from the model generation process. We do not inject or edit an existing correct model-or human-written solution. This is meant to closely approximate the response distribution that verifiers will see ``in the wild'', as they are applied in frontier math settings.
    \item \textbf{Annotation process.} We employ a \textit{strict view} of response grading: Any step that contains a mistakes or is derived from a previous mistake is considered incorrect, i.e., we do not employ ``Error Carried Forward'' grading. This is inspired by competitive math settings, the entire solution must be correct to receive full points.
\end{itemize}
Based on this philosophy, we create Hard2Verify, as we describe in detail next.

\subsection{Curating hard questions}\label{sec:setup:questions}
We construct our benchmark by collecting problem statements and official solutions \((Q, A_{\text{official}})\) from leading math competitions including the \href{https://www.imo-official.org/year_info.aspx?year=2025}{IMO}, \href{http://maa.org/putnam/}{Putnam}, and \href{https://olympiads.hbcse.tifr.res.in/mathematical-olympiad-2024-2025/}{INMO}; We provide a full list of sources in~\Cref{Tab:olympiad_stats}.
We focus question curation on recent (2024 and beyond) Olympiad-level math competitions. For each Olympiad, we parse the official PDFs using \href{https://mathpix.com/}{MathPix} and extract all content in \LaTeX{} to preserve mathematical typography and ensure stable equation rendering. We exclude image-dependent problems and only keep questions that could be solved using textual information. The resulting question set comprises 80 frontier-level problems from 10 distinct Olympiads.

\subsection{Response generation}\label{sec:setup:responses}
Using our curated question pool, we sample responses from three frontier LLMs: GPT-5 (with high reasoning), Gemini 2.5 Pro, and Claude Sonnet 4 (Thinking). We employ a standardized prompt (\Cref{prompt:generation}), instructing models to produce exam-style, stepwise proofs that mirror how an Olympiad participant would structure a solution. We use the same prompt template and decoding settings across models and disable access to external tools, like web search or code interpreters. Each model produces a single solution per problem, which we record for downstream evaluation. These samples are challenging; for example, Gemini 2.5 Pro takes up to 15 minutes to return a solution via API access. After curating all model responses to all questions, we filter out responses with undesirable qualities, such as a small number of long, dense steps or responses with degenerate outputs. This leaves us with a compact but high quality set of 200 responses. 

\subsection{Ensuring high-quality annotations}\label{sec:setup:annotations}
After sampling responses to our curated questions, human annotators meticulously annotate each model solution step-by-step. We partnered with \turingredact, a research accelerator. \turing employs mathematical experts, with a super-majority of our annotators having an advanced graduate level education in mathematics. To ensure consistent and high quality evaluations, we provided comprehensive annotation instructions as well official solutions $A_{\text{official}}$ as references. Annotation began with a multi-round pilot study, where we hand-annotated three model responses, then worked together with annotators to review samples, solicited feedback from annotators, and finetuned evaluation instructions accordingly. We then performed annotations in batches of samples, performing spot-checks of samples as they became available. This is in addition to internal processes at \turing, which include initial human annotation and three rounds of human review, where annotations were reviewed for correctness and guideline alignment. Overall, \textit{this process represents over 500 hours of manual human labor}. See~\Cref{app:annotator} for more annotation details.

\begin{figure}[t!]
    \centering
    \includegraphics[width=0.85\linewidth]{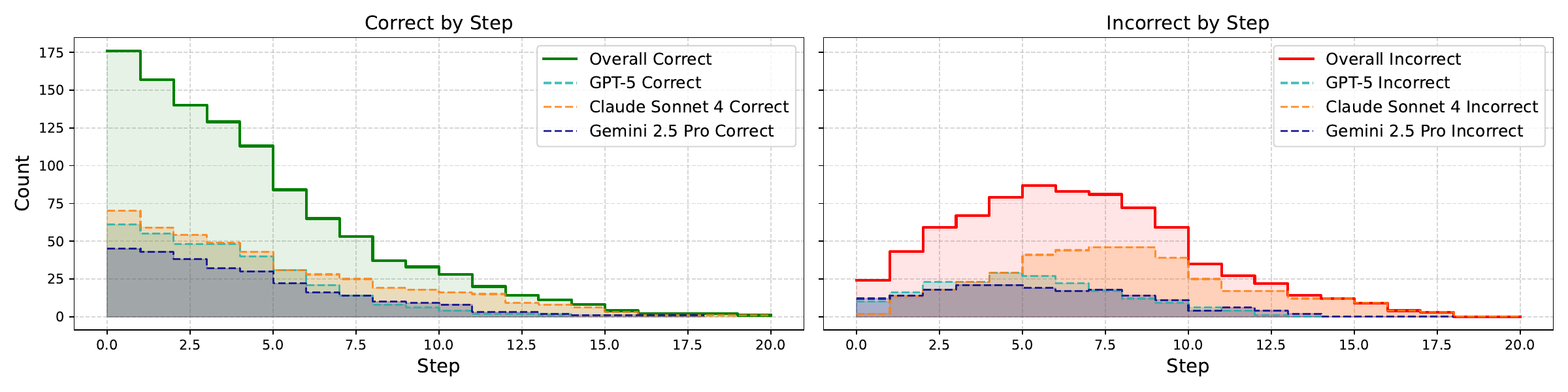}
    \caption{Count of correct (left) and incorrect (right) labels by model solution step. Models tend to begin solutions correctly, but typically start to get derailed after a few steps.}
    \label{fig:label_by_step}
    \vspace{-1\baselineskip}
\end{figure}

\subsection{Overall dataset statistics.}
Our annotation process yields 1,860 unique model steps annotated across 200 model solutions. 58\% (1,080/1,860) steps are labeled correct, while the remaining 780 are labeled incorrect.~\Cref{fig:resp_breakdown} shows how models perform on a step-level and problem level. We consider a model response correct if \textit{all} steps in the solution are graded correct by humans. Claude Sonnet 4 takes the most steps but gets the least percentage of steps correct, whereas GPT-5 and Gemini 2.5 Pro are the best performing model in terms of step-level accuracy. However, at the response level, GPT-5 outperforms Gemini 2.5 Pro by larger margins. Claude Sonnet 4, while achieving over 50\% step-level accuracy, fails to string correct steps together, only producing 4 entirely correct solutions out of 72.~\Cref{fig:label_by_step} visualizes how errors appear as a function of steps, with all three models following similar trends: Errors tend to occur in the middle of model solutions appearing after a few solution steps.

\subsection{Evaluation tasks}\label{sec:setup:tasks}
Our step-level annotations enable us to benchmark verifiers on three distinct tasks: (1) Step-level correctness (\steplevel), (2) Response-level correctness (\resplevel), and (3) First error identification (\errorid). The \steplevel task corresponds to the setup in~\citet{song2025prmbench}, whereas the \errorid tasks corresponds to that of~\citet{zheng2024processbench}. As we show in~\Cref{sec:exps}, both tasks are challenging settings for current verifiers. We provide our evaluation prompts in~\Cref{prompt:evaluation}.

\textbf{\steplevel}. Here, the verifier is tasked with determining the correctness of each step. Generative verifiers are prompted to output a binary yes/no label for each step, whereas PRM step-level scores are converted to binary labels via a fixed threshold. 

\textbf{\resplevel}. We also consider a response outcome task derived from \steplevel labels and predictions. This task reflects a strict grading of open-ended math problems: For a question to be correct, \textit{all} steps in the solution must be deemed correct. Therefore, if \textit{any} step in the solution is incorrect, we consider the solution wrong\footnote{
    Because we are concerned with ensuring completely correct responses, we apply this procedure to \textit{all} responses/questions in \hardtoverify, including non-open-ended questions.
}. From human labels, we create a single overall response-level correctness label. Likewise, given step-level predictions from a verifier, we create a response-level prediction. Note that this setting is more forgiving than the \steplevel setup: Exact step labels need not match exactly for a verifier to agree with a human at the response level. 

\textbf{\errorid}. Here, the verifier is prompted to output the first step which contains a mistake in the model solution, if present. If no error is present,  a generative verifier may output step $-1$, corresponding to ``No error''. For generative verifiers, we note that first error step can also be derived from \steplevel labels, similar to the \resplevel setting. Following ProcessBench~\citep{zheng2024processbench}, we prompt the verifier to output the step index directly. This allows us to more directly compare verifiers between benchmarks; we quantify performance differences between the direct prompting approach and converting from step-level labels in~\Cref{sec:exps:main_results}. For PRMs, we select the first step below the correctness threshold.
\begin{table}[t!]
\caption{Main evaluation results on Hard2Verify across our three evaluation tasks (\Cref{sec:setup:tasks}). We report Balanced Accuracy and Balanced F1 Score. \textbf{Best} and \underline{second-best} scores in each category marked.}
\label{tab:main_results}
\resizebox{\textwidth}{!}{%
\begin{tabular}{lcccccc}
\toprule
 & \multicolumn{2}{c}{\steplevel} & \multicolumn{2}{c}{\resplevel} & \multicolumn{2}{c}{\errorid} \\
 & Bal. Accuracy & Bal. F1 & Bal. Accuracy & Bal. F1 & Bal. Accuracy & Bal. F1 \\ 
\midrule
\rowcolor{lightgray!50}\multicolumn{7}{c}{\textit{\textbf{Generative Critics}, proprietary models}} \\
\midrule
GPT-5 & \textbf{86.53} & \textbf{85.83} & \textbf{89.69} & \textbf{89.52} & \textbf{70.61} & \textbf{69.72} \\ 
Gemini 2.5 Pro & \underline{83.37} & \underline{83.09} & \underline{85.73} & \underline{85.46} & 52.46 & 52.46 \\ 
Claude Sonnet 4 & 70.61 & 60.37 & 78.24 & 73.44 & 53.45 & 39.30 \\ 
GPT-5-Mini & 81.06 & 78.73 & 81.93 & 81.92 & 65.96 & \underline{60.04} \\ 
o3 & 78.70 & 75.29 & 83.21 & 82.58 & 60.32 & 57.31 \\ 
o4-Mini & 74.90 & 68.09 & 83.94 & 81.71 & \underline{67.31} & 57.62 \\ 
GPT-4.1 & 56.17 & 24.66 & 58.94 & 33.55 & 52.44 & 21.29 \\ 
\midrule
\rowcolor{lightgray!50}\multicolumn{7}{c}{\textit{\textbf{Generative Critics}, large ($\geq70$B) models}} \\
\midrule
Kimi K2 & 61.79 & 42.83 & 65.34 & 51.66 & 49.10 & 31.40 \\ 
DeepSeek-R1 & 68.92 & 62.30 & 73.95 & 72.75 & 54.23 & 45.35 \\
Qwen3-235B-A22B & \underline{72.55} & \underline{64.03} & \underline{79.42} & \underline{77.87} & \underline{60.90} & \underline{50.78} \\ 
Qwen3-Next-80B-A3B & 67.91 & 54.69 & 75.08 & 68.31 & 58.29 & 43.05 \\ 
Qwen2.5-72B-Instruct & 56.01 & 26.36 & 61.06 & 46.89 & 26.49 & 16.38 \\ 
GLM-4.5-Air & 57.40 & 29.40 & 61.78 & 41.00 & 41.97 & 17.81 \\ 
gpt-oss-120B & \textbf{78.10} & \textbf{74.64} & \textbf{83.92} & \textbf{83.71} & \textbf{63.97} & \textbf{60.64} \\ 
Llama-3.3-70B-Instruct & 54.28 & 18.37 & 57.04 & 28.16 & 49.44 & 2.50 \\ 
\midrule
\rowcolor{lightgray!50}\multicolumn{7}{c}{\textit{\textbf{Generative Critics}, small/medium ($<70$B) models}} \\
\midrule
Qwen3-32B & 63.99 & 51.77 & 67.86 & 63.16 & 51.96 & 26.83 \\ 
Qwen3-30B-A3B & \underline{70.71} & \underline{61.91} & 73.79 & 71.02 & \underline{58.83} & \textbf{50.51} \\ 
ByteDance Seed-OSS-36B & 66.79 & 53.09 & 72.54 & 63.88 & \textbf{59.24} & 45.18 \\ 
gpt-oss-20B & \textbf{75.18} & \textbf{70.93} & \textbf{83.85} & \textbf{83.32} & 46.13 & \underline{45.28} \\ 
Qwen3-14B & 65.48 & 52.91 & 74.59 & 70.12 & 53.69 & 37.33 \\ 
Qwen3-8B & 65.26 & 53.51 & \underline{77.61} & \underline{72.45} & 45.92 & 34.26 \\ 
Qwen2.5-14B-Instruct & 60.45 & 47.59 & 63.40 & 63.23 & 43.47 & 18.86 \\ 
Qwen2.5-7B-Instruct & 48.82 & 22.84 & 55.67 & 44.18 & 29.75 & 15.96 \\ 
\midrule
\rowcolor{lightgray!50}\multicolumn{7}{c}{\textit{\textbf{Process Reward Models}, open-source models}} \\
\midrule
Qwen2.5-Math-PRM-72B & 55.82 & 35.50 & \textbf{66.80} & \textbf{64.91} & \underline{41.80} & \textbf{37.28} \\ 
Qwen2.5-Math-PRM-7B & \underline{57.56} & \underline{42.37} & \underline{63.08} & \underline{57.57} & 35.03 & \underline{32.50} \\ 
Skywork-PRM-7B & 38.52 & 34.12 & 56.77 & 29.81 & 11.56 & 8.36 \\ 
Skywork-PRM-1.5B & 40.81 & 12.94 & 52.46 & 20.89 & 8.62 & 7.48 \\ 
ReasonFlux-PRM-7B & 53.09 & 22.40 & 55.89 & 53.82 & \textbf{42.48} & 28.71 \\ 
UniversalPRM-7B & \textbf{64.17} & \textbf{60.27} & 54.74 & 41.46 & 26.08 & 25.97 \\ 
\bottomrule
\end{tabular}%
\vspace{-1\baselineskip}
}
\end{table}

\begin{table}[t!]
\caption{Comparison of \errorid performance using two prompting approaches, with $\Delta = \steplevel - \errorid$. The \errorid prompt directly asks the verifier to identify the first step with an error, as in ProcessBench~\citep{zheng2024processbench}. The \steplevel prompt asks the verifier to produce a correctness label for each step, from which the first step with an error is derived. Balanced Accuracy tends to improve with the \steplevel prompt, but performance variation in Balanced F1 tends to be mixed.}
\label{tab:error_id}
\resizebox{\textwidth}{!}{%
\begin{tabular}{lcccccc}
 \toprule
& \errorid Prompt & \steplevel Prompt & \multirow{2}{*}{$\Delta_{\text{Bal. Acc}}$} & \errorid Prompt & \steplevel Prompt & \multirow{2}{*}{$\Delta_{\text{Bal. F1}}$}  \\
& Bal. Accuracy & Bal. Accuracy & & Bal. F1 & Bal. F1 & \\
\midrule
GPT-5 & 70.61 & 76.72 & \textcolor{ForestGreen}{+6.11}  & 69.72 & 75.66 & \textcolor{ForestGreen}{+5.94} \\ 
gpt-oss-120B & 63.97 & 69.68 & \textcolor{ForestGreen}{+5.71}  & 60.64 & 64.81 & \textcolor{ForestGreen}{+4.17} \\ 
GPT-5-Mini & 65.96 & 66.43 & \textcolor{ForestGreen}{+0.47}  & 60.04 & 63.25 & \textcolor{ForestGreen}{+3.21} \\ 
o4-Mini & 67.31 & 67.16 & \textcolor{red}{-0.15}  & 57.62 & 53.35 & \textcolor{red}{-4.27} \\ 
o3 & 60.32 & 68.02 & \textcolor{ForestGreen}{+7.70}  & 57.31 & 60.61 & \textcolor{ForestGreen}{+3.30} \\ 
Gemini 2.5 Pro & 52.46 & 66.11 & \textcolor{ForestGreen}{+13.65}  & 52.46 & 62.78 & \textcolor{ForestGreen}{+10.32} \\
Qwen3-235B-A22B & 60.90 & 65.17 & \textcolor{ForestGreen}{+4.27}  & 50.78 & 55.35 & \textcolor{ForestGreen}{+4.57} \\ 
Qwen3-30B-A3B & 58.83 & 60.19 & \textcolor{ForestGreen}{+1.36}  & 50.51 & 47.25 & \textcolor{red}{-3.26} \\ 
DeepSeek-R1 & 54.23 & 61.53 & \textcolor{ForestGreen}{+7.30}  & 45.35 & 52.02 & \textcolor{ForestGreen}{+6.67} \\ 
gpt-oss-20B & 46.13 & 66.44 & \textcolor{ForestGreen}{+20.31}  & 45.28 & 57.75 & \textcolor{ForestGreen}{+12.47} \\ 
ByteDance Seed-OSS-36B & 59.24 & 58.94 & \textcolor{red}{-0.30}  & 45.18 & 33.55 & \textcolor{red}{-11.63} \\ 
Qwen3-Next-80B-A3B & 58.29 & 63.37 & \textcolor{ForestGreen}{+5.08}  & 43.05 & 44.85 & \textcolor{ForestGreen}{+1.80} \\ 
Claude Sonnet 4 & 53.45 & 60.83 & \textcolor{ForestGreen}{+7.38}  & 39.30 & 38.59 & \textcolor{red}{-0.71} \\ 
Qwen3-14B & 53.69 & 56.56 & \textcolor{ForestGreen}{+2.87}  & 37.33 & 33.25 & \textcolor{red}{-4.08} \\ 
Qwen3-8B & 45.92 & 57.35 & \textcolor{ForestGreen}{+11.43}  & 34.26 & 29.09 & \textcolor{red}{-5.17} \\ 
Kimi K2 & 49.10 & 54.26 & \textcolor{ForestGreen}{+5.16}  & 31.40 & 23.33 & \textcolor{red}{-8.07} \\ 
Qwen3-32B & 51.96 & 52.35 & \textcolor{ForestGreen}{+0.39}  & 26.83 & 31.09 & \textcolor{ForestGreen}{+4.26} \\ 
GPT-4.1 & 52.44 & 51.97 & \textcolor{red}{-0.47}  & 21.29 & 11.89 & \textcolor{red}{-9.40} \\ 
Qwen2.5-14B-Instruct & 43.47 & 40.30 & \textcolor{red}{-3.17}  & 18.86 & 23.04 & \textcolor{ForestGreen}{+4.18} \\
GLM-4.5-Air & 41.97 & 53.24 & \textcolor{ForestGreen}{+11.27}  & 17.81 & 16.25 & \textcolor{red}{-1.56} \\ 
Qwen2.5-72B-Instruct & 26.49 & 48.09 & \textcolor{ForestGreen}{+21.60}  & 16.38 & 10.72 & \textcolor{red}{-5.66} \\ 
Qwen2.5-7B-Instruct & 29.75 & 43.01 & \textcolor{ForestGreen}{+13.26}  & 15.96 & 9.53 & \textcolor{red}{-6.43} \\ 
Llama-3.3-70B-Instruct & 49.44 & 50.71 & \textcolor{ForestGreen}{+1.27}  & 2.50 & 7.31 & \textcolor{ForestGreen}{+4.81} \\ 
\bottomrule
\end{tabular}%
}
\end{table}

\section{Experiments}\label{sec:exps}
\subsection{Evaluation Metrics}
Let TPR and TNR denote the True Positive Rate and True Negative Rate, i.e., verifier accuracies on the correct and incorrect samples, respectively. We define \textit{Balanced Accuracy} as the mean of TPR and TNR and \textit{Balanced F1 Score} as the harmonic mean of TPR and TNR\footnote{
    This is equivalent to the ``F1 Score'' used by ProcessBench, which differs from the typically used F1 Score by using TNR instead of precision. To avoid confusion, we denote this metric Balanced F1 Score.
}:
\begin{align}
    \text{Balanced F1 Score} = \frac{2\ \text{TPR}\cdot\text{TNR}}{\text{TPR} + \text{TNR}},
\end{align}
We report Balanced Accuracy and Balanced F1 Score for all tasks. The ground-truth labels and model predictions vary based on task. For \steplevel, we aggregate all steps and all verifier predictions across all responses, whereas for \resplevel and \errorid, we compute metrics at the response level. These metrics quantify verifier behavior in terms of correctly identifying mistakes versus correct answers. Balanced Accuracy and Balanced F1 both serve as aggregate measures: the former reflects average performance across both modes, while the latter penalizes imbalanced performance. An ideal verifier scores highly on both.

\subsection{Evaluated models}
We select a variety of PRMs and generative models prompted as step-level critics. For prompted generative critics, we test a closed-source frontier models as well as large ($\geq70$B) and small-medium ($<$70B) open-weight models. We evaluate all reasoning models at their maximum provided reasoning level (e.g., ``high'' for GPT-5), using suggested sampling parameters for various baselines. All Qwen3 models are evaluated with ``thinking on''. For instruction-tuned models, we use greedy decoding. For all models, we set the maximum number of output tokens to be 32K. The full set of models is enumerated in~\Cref{app:baselines}. For PRMs, we select Qwen2.5-Math-PRM-\{7B,72B\}~\citep{zhang2025lessons}, Skywork-PRM-\{1.5B,7B\}~\citep{he_2024_16998085}, ReasonFlux-PRM-\{1.5B,7B\}~\citep{zou2025reasonflux}, and UniversalPRM~\citep{tan2025aurora}. We tune PRM thresholds following~\citet{zheng2024processbench}; See~\Cref{app:prm:threshold} for more details.

\subsection{Main evaluation results}\label{sec:exps:main_results}
\Cref{tab:main_results} presents our main results, with detailed results presented in~\Cref{app:additional_results}. Among proprietary models GPT-5 stands out in its overall ability across all three tasks, able to accurately provide step-level correctness labels and identify the first error in reasoning. Gemini 2.5 Pro follows closely for step-level identification, but lags in error identification. Finally, Claude Sonnet 4 with Thinking enabled lags OpenAI models and Gemini 2.5 Pro, failing to match even reasoning models from previous generations, like o3 and o4-mini.

Among larger open-weight models, the gpt-oss series are clear standouts, with gpt-oss-120B roughly matching the performance of o3. Recent larger Qwen3 models and DeepSeek-R1 challenge for second place in step-level verification, but all lag in terms of error identification. Notably, Llama-3.3-70B, which performs admirably on ProcessBench (58.0 Balanced F1; \Cref{fig:teaser}) performs extremely poorly, achieving only 2.50 on Balanced F1 for the error identification task.

Among smaller models, gpt-oss-20B performs extremely well on step-level and response-level tasks, but falters in identifying errors. ByteDance Seed-OSS-36B and Qwen3-30B-A3B are the next best performers, but only ByteDance Seed-OSS-36B is able to outperform random guessing\footnote{Here, we consider a verifier that ``randomly guesses'' as one that achieves 50\% TPR and TNR, yielding Balanced Accuracy and Balanced F1 scores of 50.0.} performance in error identification. Finally, even state-of-the-art PRMs, like the Qwen2.5-Math-PRM series struggle immensely on \hardtoverify, performing significantly below random guess performance in error identification. For example, Qwen2.5-Math-PRM-72B achieves only 37.28\% Balanced F1.

\begin{figure}[t!]
    \centering
    \includegraphics[width=0.9\linewidth]{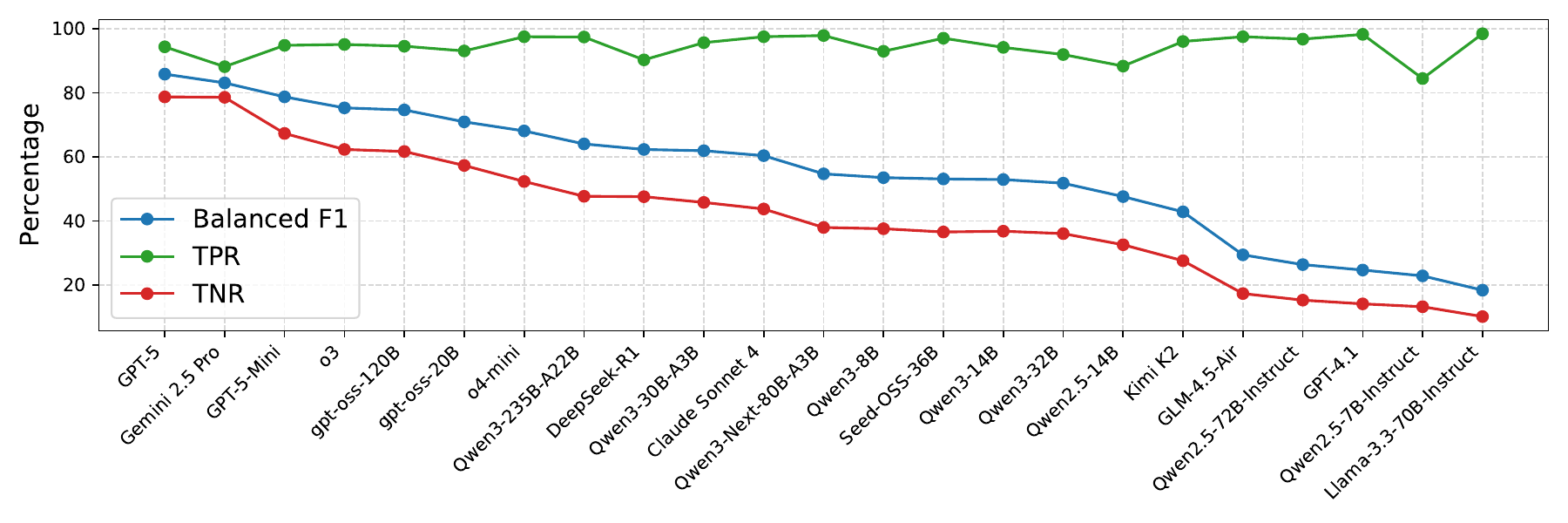}
    \caption{Weaker models are unable to find mistakes, eventually considering \textit{all} steps correct: TNR tends toward 0 while TPR tends towards 1.}
    \label{fig:tpr_tnr_plot}
    \vspace{-1\baselineskip}
\end{figure}

\paragraph{What separates strong verifiers from weak verifiers?} To provide additional insights into performance variations across different models,~\Cref{fig:tpr_tnr_plot} plots the TPR and TNR for all generative critics models, sorted in performance from strongest (left) to weakest (right) in terms of Balanced F1 Score. A clear trend emerges: Verifier performance degrades because TNR drops quickly to near 0, while TPR rises gradually to almost 1. This indicates that almost all steps are labeled as correct, revealing that \textit{weaker verifiers cannot catch errors}. Notably, the order of models from left to right approximately correlates with mathematical \textit{generation} ability, i.e., the ability to solve extremely difficult math problems. As such, this may indicate that a baseline level of solving ability is a necessary prerequisite for verification.~\Cref{app:additional_results} shows this trend holds similarly for \resplevel and \errorid tasks.

\paragraph{To identify errors, how should generative verifiers be prompted?} Our \errorid task adopts the setup of ProcessBench~\citep{zheng2024processbench}, which directly prompts the verifier to output the index of the first step with an error. However, the index of the first error can also be derived from step-level correctness labels, like those produced in the \steplevel task. In~\Cref{tab:error_id}, we compare the performance under the \errorid Prompt and \steplevel Prompt. 

Surprisingly, directly prompting for the given task may not yield the best performance: In terms of Balanced F1, performance across models is mixed, with some models exhibiting very small performance changes and others exhibiting significant changes. For example, switching from direct to step-level prompting significantly degrades performance for ByteDance Seed-OSS-36B from 45.18 to 33.55, while boosting performance for Gemini 2.5 Pro from 52.46 to 62.78. Overall, we find that more capable models, like GPT-5 and Gemini 2.5 Pro, benefit the most from switching to deriving first identified error from \steplevel outputs. 

We hypothesize that requiring step-by-step annotations requires models to more closely inspect each step, allowing for better error identification. This generally tends to benefit stronger models more capable of performing step-level verification well, but can hurt insufficiently capable models. As the change in performance is mixed across models, we advise practitioners to determine the better prompting setup on a per-verifier basis.

\section{Additional Analysis}

\begin{figure}[t!]
    \vspace{-1\baselineskip}
    \centering
    \includegraphics[width=0.66\linewidth]{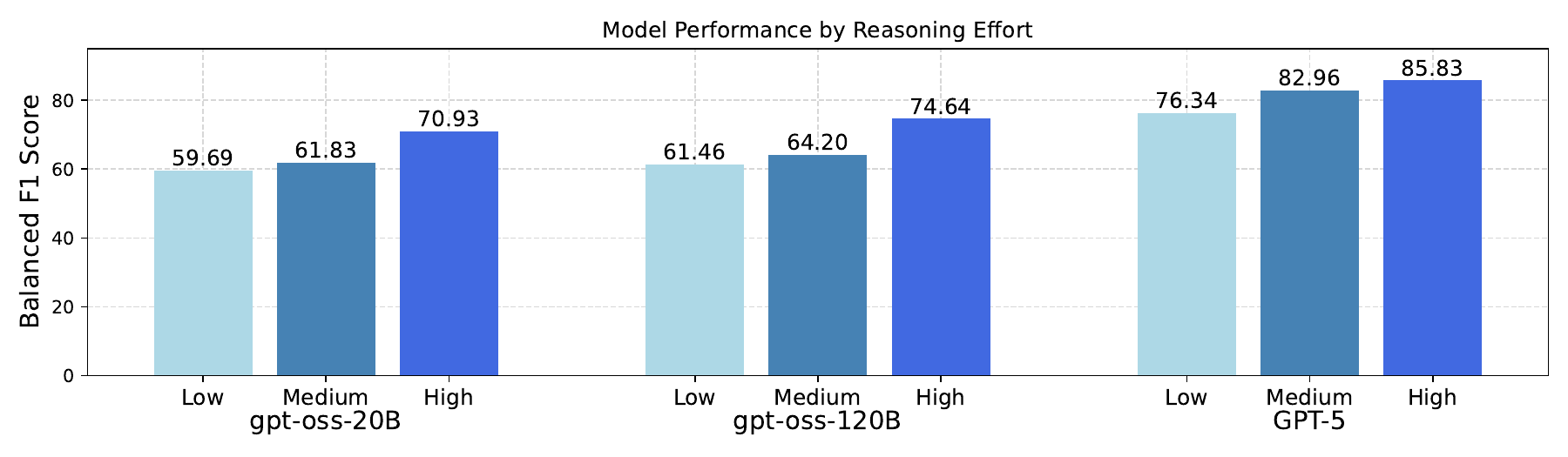}
    \includegraphics[width=0.66\linewidth]{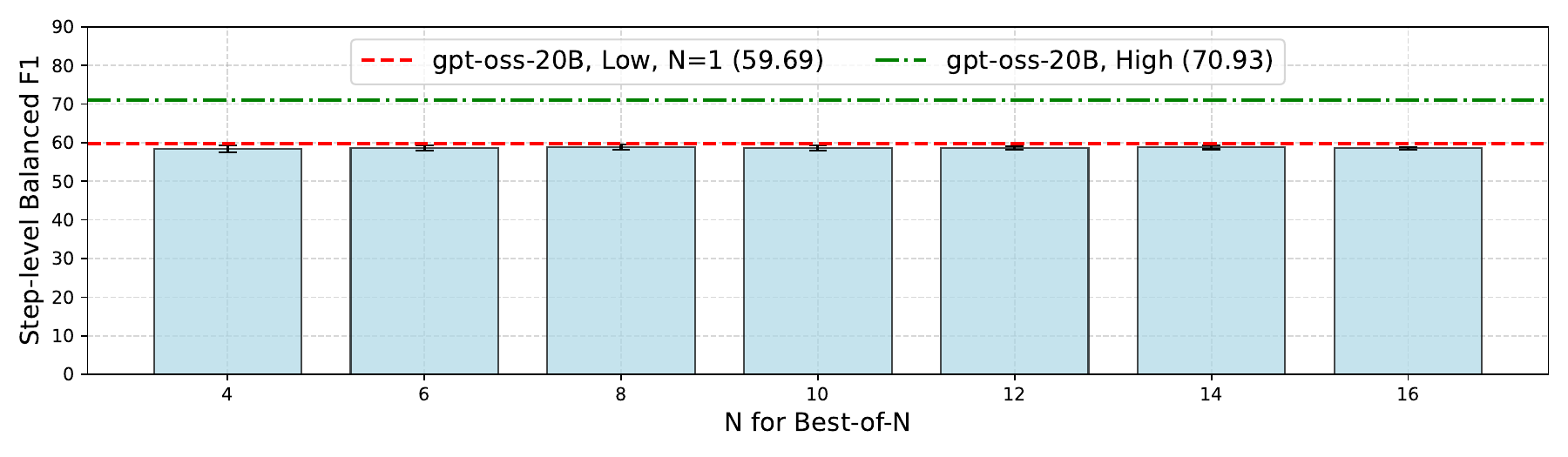}
    \caption{\textbf{Top}: Scaling inference-time compute sequentially leads to higher performance in GPT-5 and gpt-oss models, with large gains for gpt-oss-20B (59.69$\rightarrow$70.93) and 120B (61.46 $\rightarrow$ 74.64) in terms of step-level Balanced F1. 
    \textbf{Bottom}: Parallel decoding has little effect on step-level F1 performance for gpt-oss-20B, failing to bridge the gap vs. gpt-oss-20B at high-reasoning effort.}
    \label{fig:seq_scaling}
\end{figure}
\subsection{How should we scale verifier inference-time compute?}\label{sec:analysis:inf_time}
Here, we experiment with scaling verifier inference-time compute along via sequential and parallel approaches. We find sequential scaling brings substantive gains, whereas parallel scaling does not.

\textbf{Sequential inference-time compute scaling.} Here we explore scaling inference-time compute sequentially by letting the verifier output more tokens when verifying, focusing on the \steplevel task. We use gpt-oss-20B, gpt-oss-120B, and GPT-5, which all have three distinct reasoning levels: low, medium, and high. In~\Cref{fig:seq_scaling} (top), we plot Balanced F1. Affording the verifier to generate more ``thinking'' tokens at inference time generally improves performance, with gpt-oss-120B improving the most from low (61.46) to high (74.64) and gpt-oss-20B likewise improving significantly. Gains for GPT-5 are smaller compared to gpt-oss models, but still significant, with 12.3\% relative improvement from low to high. 

\textbf{Parallel inference-time compute scaling.} Here, we attempt to match the performance of gpt-oss-20B at high reasoning effort by sampling $N$ outputs in parallel from gpt-oss-20B at low reasoning effort. To do so, we sample $32$ responses per sample from gpt-oss-20B and simulate best of $N$ from $N = 4,\ldots,16$ via bootstrap sampling. Concretely, for each $N$, we sample $N$ responses from the $32$ with replacement, and aggregate predicted step-level labels via majority vote, breaking ties arbitrarily. To reduce variance, for each value of $N$, we repeat this process for 10 trials, and report mean and standard deviation across the 10 trials in~\Cref{fig:seq_scaling} (bottom).
We also plot the baseline gpt-oss-20B performance at low and high reasoning efforts. Surprisingly, Best-of-$N$ does not meaningfully improve over sampling $1$ response as $N$ increases. An intuitive explanation for this phenomenon is that step-level verification is inherently a sequential task: Each step must be processed one-after-another. As such, affording the verifier more time to ``think'' about each step is more effective than sampling multiple ``rushed'' judgments.

\begin{figure}[b!]
    \centering
    \includegraphics[width=0.75\linewidth]{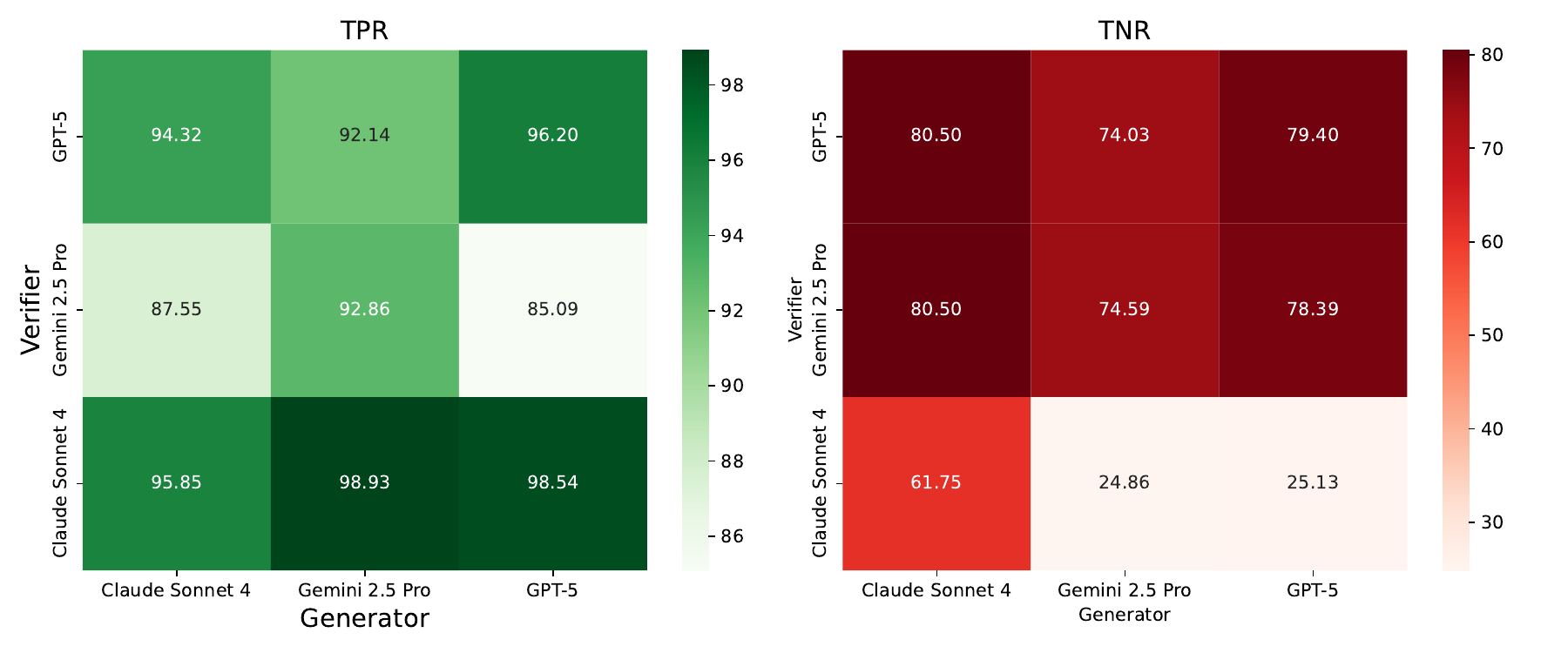}
    \caption{Verifier TPR and TNR based on response generator model. For strong verifiers (GPT-5, Gemini 2.5 Pro), TPR varies based on generator, with GPT-5 exhibiting the most stable performance on a per-generator basis. Claude Sonnet 4 generates the easiest to catch mistakes across all verifiers, whereas Gemini 2.5 Pro produces the hardest to catch mistakes, as measured by TNR.}
    \label{fig:own_resp}
\end{figure}
\subsection{How do verifiers verify their own responses?}\label{sec:analysis:self_verif}
 
We investigate the dynamics of self-verification, focusing on GPT-5, Gemini 2.5 Pro, and Claude Sonnet 4 as verifiers.~\Cref{fig:own_resp} plots the step-level TPR and TNR performance based on response generator. The results notably depend on verifier strength:~\Cref{tab:main_results} shows that GPT-5 and Gemini 2.5 Pro are the top two performers, whereas Claude Sonnet 4 is a relatively weak proprietary verifier. 
 
 We find that GPT-5 and Gemini 2.5 Pro as verifiers are more likely to consider a correct self-generated response as correct, as measured by TPR. Of the two, GPT-5 exhibits the least variation in TPR across models, while Gemini 2.5 Pro performance drops from 92.86 TPR on own-generated responses to as low as 85.09 TPR for GPT-5-generated responses. Claude Sonnet 4, on the other hand, overwhelmingly assigns ``Correct'' as a label, leading to high TPRs regardless of generator. 

 Across all threee models, it is easier to identify errors from the weaker model (Claude Sonnet 4) than it is to identify errors from the stronger models. This result is consistent with recent work~\citep{zhou2025variation} studying verification, which finds weaker generators produce easier to catch errors. Interestingly, both GPT-5 and Gemini 2.5 Pro struggle have the lowest TNR on responses from Gemini 2.5 Pro, showing that GPT-5 is more reliable in self-critique than Gemini 2.5 Pro is. The fact that Gemini 2.5 Pro has the lowest TNR on self-generated responses is consistent with recent work analyzing self-reflection~\citep{stechly2023gpt,stechly2024self,huang2023large}, where LLMs were shown to have difficulties correcting their own mistakes in challenging reasoning settings. In contrast, Claude Sonnet 4 as a relatively weaker verifier cannot identify errors in stronger model responses.

\begin{figure}[t!]
    \centering
    \includegraphics[width=0.7\linewidth]{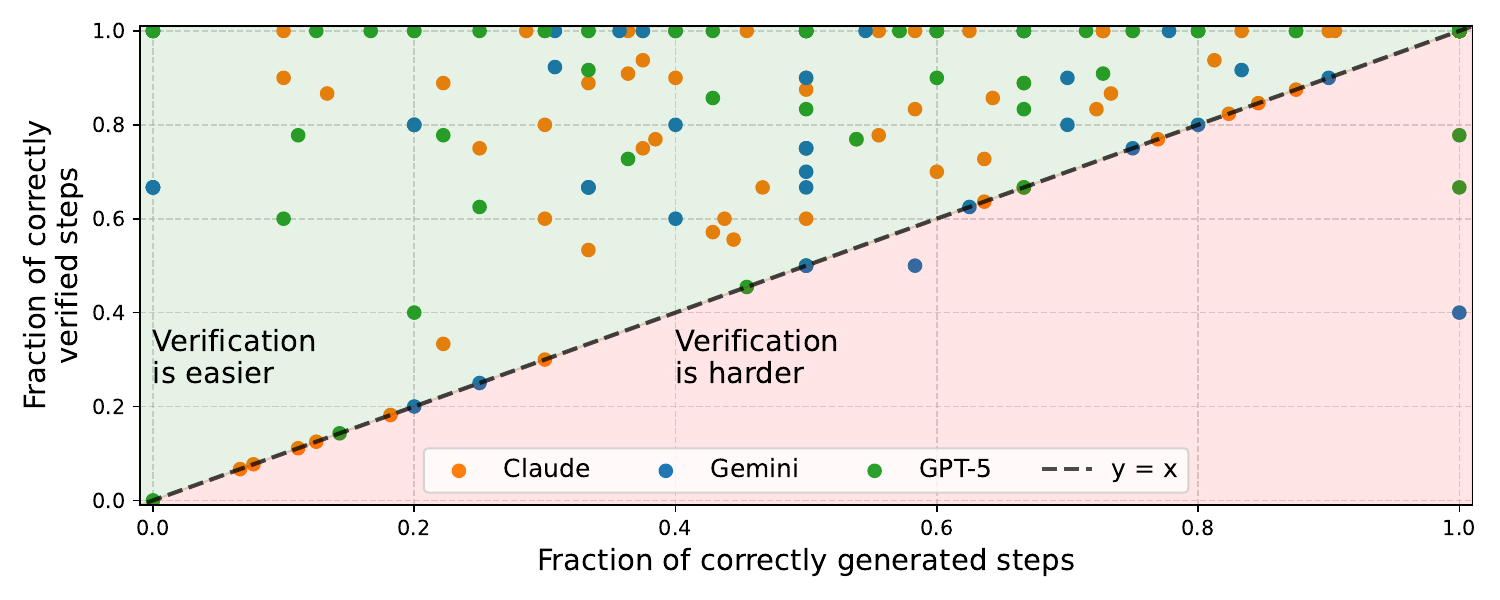}
    \caption{Each generators evaluates self-produced responses, and the fraction of steps correctly solved vs. fraction of steps correctly verified for a given question is plotted. In general, we find that models are more successful in catching mistakes than generating error-free responses.}
    \label{fig:gen_vs_verif}
    \vspace{-1\baselineskip}
\end{figure}
\subsection{Is verifying problems easier than solving problems?}
Here, we examine if generating a solution is easier than verifying the same solution. We split Hard2Verify into three subsets corresponding to each of the three generator models and have the generators verify their own responses. For each response, we record the fraction of correctly generated steps (``solve rate''), as deemed by human annotators, and the fraction of correctly verified steps (``verification rate''), as deemed by agreement with human annotators. In~\Cref{fig:gen_vs_verif}, we plot the verification rate against the solve rate. We observe that the verification rate is consistently higher than the solve rate across all models; Only on a few problems does the verifier have a more difficult time verifying a problem than generating the problem. This result offers some optimism for future work in verification: Because verifying a solution tends to be ``easier'' than generating the solution, verifiers may not necessarily need to be as powerful as frontier generators to reliably identify errors.

\subsection{Case study: Where do models and humans disagree?}
We inspect outputs from a relatively strong open-source verifier, ByteDance Seed-OSS-36B~\citep{seed2025seed-oss} on multiple IMO-level problems and found a recurring theme: The verifier incorrectly accepts \textit{partial or under-justified claims as correct}. We provide two concrete examples below. These mismatches reflect larger \textit{systematic} behavior in verifiers, revealed in~\Cref{sec:exps}: Current verifiers are too \textit{generous}, with TPR rate tending towards 1 and TNR tending toward 0, indicating that a vast majority steps are considered correct.

On IMO 2023 Shortlist, question A6, Gemini 2.5 Pro makes a generalized claim, but only proves the claim for a single input. Human annotators catch this mistake, noting ``\emph{The equality holds only at one point ... not a polynomial identity, so coefficients need not match}.'' Seed-OSS-36B considers this step correct without mentioning the unfounded generalization. Similarly, on IMO 2024 Shortlist, question A1, Claude Sonnet 4 as generator constructs a proof by cases by invoking Weyl's equidistribution theorem, but considers only a single case: ``\emph{if $\alpha$ is not an even integer, then $\alpha=m+\beta$ with $m$ odd and $2/3\le\beta<1$...}''. Seed-OSS-36B greenlights this step as correct, whereas human annotators find it incomplete: ``\emph{The case analysis ignores the branch where $m$ is even and $0<\beta<1/3$,...}''. Further, the theorem invocation itself is deemed under-specified: ``\emph{justification [for invoking Weyl's equidistribution theorem] should explicitly specify the estimate and the choice of $n$}''.

\section{Conclusion}
We introduce \hardtoverify, a human-annotated, step-level benchmark aimed to assess how step-level verifiers operate in frontier settings. We focus our data curation on recent open-ended math problems, sampling responses from frontier LLMs. The end result of over 500 hours of human annotation effort is a benchmark that challenges many current open-source verifiers, which are unable to match the performance of larger, proprietary models.

\section*{Acknowledgments}
For human annotation, we partnered with Turing\footnote{\url{https://www.turing.com/}}, a research accelerator that specializes in frontier data curation and annotation for AI applications.

\newpage
\bibliography{iclr2026_conference}

\begin{thebibliography}{58}
\providecommand{\natexlab}[1]{#1}
\providecommand{\url}[1]{\texttt{#1}}
\expandafter\ifx\csname urlstyle\endcsname\relax
  \providecommand{\doi}[1]{doi: #1}\else
  \providecommand{\doi}{doi: \begingroup \urlstyle{rm}\Url}\fi

\bibitem[Agarwal et~al.(2025)Agarwal, Ahmad, Ai, Altman, Applebaum, Arbus, Arora, Bai, Baker, Bao, et~al.]{gptoss_agarwal2025}
Sandhini Agarwal, Lama Ahmad, Jason Ai, Sam Altman, Andy Applebaum, Edwin Arbus, Rahul~K Arora, Yu~Bai, Bowen Baker, Haiming Bao, et~al.
\newblock gpt-oss-120b \& gpt-oss-20b model card.
\newblock \emph{arXiv preprint arXiv:2508.10925}, 2025.

\bibitem[Anthropic(2025)]{claude4}
Anthropic.
\newblock Introducing claude 4.
\newblock 2025.
\newblock URL \url{https://www.anthropic.com/news/claude-4}.

\bibitem[Cobbe et~al.(2021)Cobbe, Kosaraju, Bavarian, Chen, Jun, Kaiser, Plappert, Tworek, Hilton, Nakano, Hesse, and Schulman]{cobbe2021gsm8k}
Karl Cobbe, Vineet Kosaraju, Mohammad Bavarian, Mark Chen, Heewoo Jun, Lukasz Kaiser, Matthias Plappert, Jerry Tworek, Jacob Hilton, Reiichiro Nakano, Christopher Hesse, and John Schulman.
\newblock Training verifiers to solve math word problems.
\newblock \emph{arXiv preprint arXiv:2110.14168}, 2021.

\bibitem[Frick et~al.(2024)Frick, Li, Chen, Chiang, Angelopoulos, Jiao, Zhu, Gonzalez, and Stoica]{frick2024evaluate}
Evan Frick, Tianle Li, Connor Chen, Wei-Lin Chiang, Anastasios~N Angelopoulos, Jiantao Jiao, Banghua Zhu, Joseph~E Gonzalez, and Ion Stoica.
\newblock How to evaluate reward models for rlhf.
\newblock \emph{arXiv preprint arXiv:2410.14872}, 2024.

\bibitem[Gao et~al.(2024)Gao, Song, Yang, Cai, Miao, Dong, Li, Ma, Chen, Xu, et~al.]{gao2024omni}
Bofei Gao, Feifan Song, Zhe Yang, Zefan Cai, Yibo Miao, Qingxiu Dong, Lei Li, Chenghao Ma, Liang Chen, Runxin Xu, et~al.
\newblock Omni-math: A universal olympiad level mathematic benchmark for large language models.
\newblock \emph{arXiv preprint arXiv:2410.07985}, 2024.

\bibitem[Glazer et~al.(2024)Glazer, Erdil, Besiroglu, Chicharro, Chen, Gunning, Olsson, Denain, Ho, Santos, et~al.]{glazer2024frontiermath}
Elliot Glazer, Ege Erdil, Tamay Besiroglu, Diego Chicharro, Evan Chen, Alex Gunning, Caroline~Falkman Olsson, Jean-Stanislas Denain, Anson Ho, Emily de~Oliveira Santos, et~al.
\newblock Frontiermath: A benchmark for evaluating advanced mathematical reasoning in ai.
\newblock \emph{arXiv preprint arXiv:2411.04872}, 2024.

\bibitem[Google(2025{\natexlab{a}})]{gemini_25}
Google.
\newblock Gemini 2.5: Our most intelligent ai model.
\newblock \emph{https://blog.google/technology/google-deepmind/gemini-model-thinking-updates-march-2025/}, 2025{\natexlab{a}}.

\bibitem[Google(2025{\natexlab{b}})]{gemini_deep_think_imo}
Google.
\newblock Advanced version of gemini with deep think officially achieves gold-medal standard at the international mathematical olympiad.
\newblock \emph{https://deepmind.google/discover/blog/advanced-version-of-gemini-with-deep-think-officially-achieves-gold-medal-standard-at-the-international-mathematical-olympiad/}, 2025{\natexlab{b}}.

\bibitem[Grattafiori et~al.(2024)Grattafiori, Dubey, Jauhri, Pandey, Kadian, Al-Dahle, Letman, Mathur, Schelten, Vaughan, et~al.]{grattafiori2024llama}
Aaron Grattafiori, Abhimanyu Dubey, Abhinav Jauhri, Abhinav Pandey, Abhishek Kadian, Ahmad Al-Dahle, Aiesha Letman, Akhil Mathur, Alan Schelten, Alex Vaughan, et~al.
\newblock The llama 3 herd of models.
\newblock \emph{arXiv preprint arXiv:2407.21783}, 2024.

\bibitem[Guo et~al.(2025)Guo, Yang, Zhang, Song, Zhang, Xu, Zhu, Ma, Wang, Bi, et~al.]{guo2025deepseek}
Daya Guo, Dejian Yang, Haowei Zhang, Junxiao Song, Ruoyu Zhang, Runxin Xu, Qihao Zhu, Shirong Ma, Peiyi Wang, Xiao Bi, et~al.
\newblock Deepseek-r1: Incentivizing reasoning capability in llms via reinforcement learning.
\newblock \emph{arXiv preprint arXiv:2501.12948}, 2025.

\bibitem[He et~al.(2024{\natexlab{a}})He, Luo, Bai, Hu, Thai, Shen, Hu, Han, Huang, Zhang, et~al.]{he2024olympiadbench}
Chaoqun He, Renjie Luo, Yuzhuo Bai, Shengding Hu, Zhen~Leng Thai, Junhao Shen, Jinyi Hu, Xu~Han, Yujie Huang, Yuxiang Zhang, et~al.
\newblock Olympiadbench: A challenging benchmark for promoting agi with olympiad-level bilingual multimodal scientific problems.
\newblock \emph{arXiv preprint arXiv:2402.14008}, 2024{\natexlab{a}}.

\bibitem[He et~al.(2024{\natexlab{b}})He, Wei, Yan, Liu, Wang, Gan, Tu, Liu, Zeng, Wang, Wang, Li, Zhang, Xu, An, Liu, and Zhou]{he_2024_16998085}
Jujie He, Tianwen Wei, Rui Yan, Jiacai Liu, Chaojie Wang, Yimeng Gan, Shiwen Tu, Chris~Yuhao Liu, Liang Zeng, Xiaokun Wang, Boyang Wang, Yongcong Li, Fuxiang Zhang, Jiacheng Xu, Bo~An, Yang Liu, and Yahui Zhou.
\newblock Skywork-o1 open series, November 2024{\natexlab{b}}.
\newblock URL \url{https://doi.org/10.5281/zenodo.16998085}.

\bibitem[Hendrycks et~al.(2020)Hendrycks, Burns, Basart, Zou, Mazeika, Song, and Steinhardt]{hendrycks2020measuring}
Dan Hendrycks, Collin Burns, Steven Basart, Andy Zou, Mantas Mazeika, Dawn Song, and Jacob Steinhardt.
\newblock Measuring massive multitask language understanding.
\newblock \emph{arXiv preprint arXiv:2009.03300}, 2020.

\bibitem[Hendrycks et~al.(2021)Hendrycks, Burns, Kadavath, Arora, Basart, Tang, Song, and Steinhardt]{hendrycks2021measuring}
Dan Hendrycks, Collin Burns, Saurav Kadavath, Akul Arora, Steven Basart, Eric Tang, Dawn Song, and Jacob Steinhardt.
\newblock Measuring mathematical problem solving with the math dataset.
\newblock \emph{arXiv preprint arXiv:2103.03874}, 2021.

\bibitem[Huang et~al.(2023)Huang, Chen, Mishra, Zheng, Yu, Song, and Zhou]{huang2023large}
Jie Huang, Xinyun Chen, Swaroop Mishra, Huaixiu~Steven Zheng, Adams~Wei Yu, Xinying Song, and Denny Zhou.
\newblock Large language models cannot self-correct reasoning yet.
\newblock \emph{arXiv preprint arXiv:2310.01798}, 2023.

\bibitem[Huang \& Yang(2025)Huang and Yang]{huang2025gemini}
Yichen Huang and Lin~F Yang.
\newblock Gemini 2.5 pro capable of winning gold at imo 2025.
\newblock \emph{arXiv preprint arXiv:2507.15855}, 2025.

\bibitem[Ke et~al.(2025)Ke, Jiao, Ming, Nguyen, Xu, Long, Li, Qin, Wang, Savarese, et~al.]{ke2025survey}
Zixuan Ke, Fangkai Jiao, Yifei Ming, Xuan-Phi Nguyen, Austin Xu, Do~Xuan Long, Minzhi Li, Chengwei Qin, Peifeng Wang, Silvio Savarese, et~al.
\newblock A survey of frontiers in llm reasoning: Inference scaling, learning to reason, and agentic systems.
\newblock \emph{arXiv preprint arXiv:2504.09037}, 2025.

\bibitem[Kwon et~al.(2023)Kwon, Li, Zhuang, Sheng, Zheng, Yu, Gonzalez, Zhang, and Stoica]{kwon2023efficient}
Woosuk Kwon, Zhuohan Li, Siyuan Zhuang, Ying Sheng, Lianmin Zheng, Cody~Hao Yu, Joseph~E. Gonzalez, Hao Zhang, and Ion Stoica.
\newblock Efficient memory management for large language model serving with pagedattention.
\newblock In \emph{Proceedings of the ACM SIGOPS 29th Symposium on Operating Systems Principles}, 2023.

\bibitem[Lambert et~al.(2024)Lambert, Morrison, Pyatkin, Huang, Ivison, Brahman, Miranda, Liu, Dziri, Lyu, et~al.]{lambert2024tulu}
Nathan Lambert, Jacob Morrison, Valentina Pyatkin, Shengyi Huang, Hamish Ivison, Faeze Brahman, Lester James~V Miranda, Alisa Liu, Nouha Dziri, Shane Lyu, et~al.
\newblock Tulu 3: Pushing frontiers in open language model post-training.
\newblock \emph{arXiv preprint arXiv:2411.15124}, 2024.

\bibitem[Lifshitz et~al.(2025)Lifshitz, McIlraith, and Du]{lifshitz2025multi}
Shalev Lifshitz, Sheila~A McIlraith, and Yilun Du.
\newblock Multi-agent verification: Scaling test-time compute with multiple verifiers.
\newblock \emph{arXiv preprint arXiv:2502.20379}, 2025.

\bibitem[Lightman et~al.(2023)Lightman, Kosaraju, Burda, Edwards, Baker, Lee, Leike, Schulman, Sutskever, and Cobbe]{lightman2023let}
Hunter Lightman, Vineet Kosaraju, Yuri Burda, Harrison Edwards, Bowen Baker, Teddy Lee, Jan Leike, John Schulman, Ilya Sutskever, and Karl Cobbe.
\newblock Let's verify step by step.
\newblock In \emph{The Twelfth International Conference on Learning Representations}, 2023.

\bibitem[Liu et~al.(2025)Liu, Wang, Xu, Ma, Ruan, Li, Liu, and Wu]{liu2025inference}
Zijun Liu, Peiyi Wang, Runxin Xu, Shirong Ma, Chong Ruan, Peng Li, Yang Liu, and Yu~Wu.
\newblock Inference-time scaling for generalist reward modeling.
\newblock \emph{arXiv preprint arXiv:2504.02495}, 2025.

\bibitem[Luo et~al.(2024)Luo, Liu, Liu, Phatale, Lara, Li, Shu, Zhu, Meng, Sun, et~al.]{luo2024improve}
Liangchen Luo, Yinxiao Liu, Rosanne Liu, Samrat Phatale, Harsh Lara, Yunxuan Li, Lei Shu, Yun Zhu, Lei Meng, Jiao Sun, et~al.
\newblock Improve mathematical reasoning in language models by automated process supervision.
\newblock \emph{arXiv preprint arXiv:2406.06592}, 2024.

\bibitem[Mahan et~al.(2024)Mahan, Van~Phung, Rafailov, Blagden, Lile, Castricato, Fr{\"a}nken, Finn, and Albalak]{mahan2024generative}
Dakota Mahan, Duy Van~Phung, Rafael Rafailov, Chase Blagden, Nathan Lile, Louis Castricato, Jan-Philipp Fr{\"a}nken, Chelsea Finn, and Alon Albalak.
\newblock Generative reward models.
\newblock \emph{arXiv preprint arXiv:2410.12832}, 2024.

\bibitem[MMA(2025)]{aime}
MMA.
\newblock (american invitational mathematics examination).
\newblock \emph{https://maa.org}, 2025.

\bibitem[OpenAI(2025{\natexlab{a}})]{gpt5}
OpenAI.
\newblock Gpt-5 system card.
\newblock 2025{\natexlab{a}}.
\newblock URL \url{https://cdn.openai.com/gpt-5-system-card.pdf}.

\bibitem[OpenAI(2025{\natexlab{b}})]{o4mini}
OpenAI.
\newblock Openai o3 and o4-mini system card.
\newblock 2025{\natexlab{b}}.
\newblock URL \url{https://cdn.openai.com/pdf/2221c875-02dc-4789-800b-e7758f3722c1/o3-and-o4-mini-system-card.pdf}.

\bibitem[{OpenAI}(2025)]{openai_gpt41_2025}
{OpenAI}.
\newblock Introducing {GPT-4.1} in the api.
\newblock \url{https://openai.com/index/gpt-4-1/}, April 2025.
\newblock Accessed: 2025-09-25.

\bibitem[OpenAI(2025)]{openai_imo}
OpenAI.
\newblock Openai imo 2025 proofs.
\newblock \emph{https://github.com/aw31/openai-imo-2025-proofs}, 2025.

\bibitem[Phan et~al.(2025)Phan, Gatti, Han, Li, Hu, Zhang, Zhang, Shaaban, Ling, Shi, et~al.]{phan2025humanity}
Long Phan, Alice Gatti, Ziwen Han, Nathaniel Li, Josephina Hu, Hugh Zhang, Chen Bo~Calvin Zhang, Mohamed Shaaban, John Ling, Sean Shi, et~al.
\newblock Humanity's last exam.
\newblock \emph{arXiv preprint arXiv:2501.14249}, 2025.

\bibitem[Setlur et~al.(2025)Setlur, Nagpal, Fisch, Geng, Eisenstein, Agarwal, Agarwal, Berant, and Kumar]{setlur2025rewarding}
Amrith Setlur, Chirag Nagpal, Adam Fisch, Xinyang Geng, Jacob Eisenstein, Rishabh Agarwal, Alekh Agarwal, Jonathan Berant, and Aviral Kumar.
\newblock Rewarding progress: Scaling automated process verifiers for {LLM} reasoning.
\newblock In \emph{The Thirteenth International Conference on Learning Representations}, 2025.
\newblock URL \url{https://openreview.net/forum?id=A6Y7AqlzLW}.

\bibitem[Shao et~al.(2024)Shao, Wang, Zhu, Xu, Song, Bi, Zhang, Zhang, Li, Wu, et~al.]{grpo_shao2024deepseekmath}
Zhihong Shao, Peiyi Wang, Qihao Zhu, Runxin Xu, Junxiao Song, Xiao Bi, Haowei Zhang, Mingchuan Zhang, YK~Li, Yang Wu, et~al.
\newblock Deepseekmath: Pushing the limits of mathematical reasoning in open language models.
\newblock \emph{arXiv preprint arXiv:2402.03300}, 2024.

\bibitem[Snell et~al.(2024)Snell, Lee, Xu, and Kumar]{scaling_testtime_optimallysnell2024scaling}
Charlie Snell, Jaehoon Lee, Kelvin Xu, and Aviral Kumar.
\newblock Scaling llm test-time compute optimally can be more effective than scaling model parameters.
\newblock \emph{arXiv preprint arXiv:2408.03314}, 2024.

\bibitem[Song et~al.(2025)Song, Su, Qu, Zhou, and Cheng]{song2025prmbench}
Mingyang Song, Zhaochen Su, Xiaoye Qu, Jiawei Zhou, and Yu~Cheng.
\newblock Prmbench: A fine-grained and challenging benchmark for process-level reward models.
\newblock \emph{arXiv preprint arXiv:2501.03124}, 2025.

\bibitem[Stechly et~al.(2023)Stechly, Marquez, and Kambhampati]{stechly2023gpt}
Kaya Stechly, Matthew Marquez, and Subbarao Kambhampati.
\newblock Gpt-4 doesn't know it's wrong: An analysis of iterative prompting for reasoning problems.
\newblock \emph{arXiv preprint arXiv:2310.12397}, 2023.

\bibitem[Stechly et~al.(2024)Stechly, Valmeekam, and Kambhampati]{stechly2024self}
Kaya Stechly, Karthik Valmeekam, and Subbarao Kambhampati.
\newblock On the self-verification limitations of large language models on reasoning and planning tasks.
\newblock \emph{arXiv preprint arXiv:2402.08115}, 2024.

\bibitem[Tan et~al.(2024)Tan, Zhuang, Montgomery, Tang, Cuadron, Wang, Popa, and Stoica]{tan2024judgebench}
Sijun Tan, Siyuan Zhuang, Kyle Montgomery, William~Y Tang, Alejandro Cuadron, Chenguang Wang, Raluca~Ada Popa, and Ion Stoica.
\newblock Judgebench: A benchmark for evaluating llm-based judges.
\newblock \emph{arXiv preprint arXiv:2410.12784}, 2024.

\bibitem[Tan et~al.(2025)Tan, Yao, Qu, Li, Yang, Lu, Wang, Qiu, Chu, Xu, et~al.]{tan2025aurora}
Xiaoyu Tan, Tianchu Yao, Chao Qu, Bin Li, Minghao Yang, Dakuan Lu, Haozhe Wang, Xihe Qiu, Wei Chu, Yinghui Xu, et~al.
\newblock Aurora: Automated training framework of universal process reward models via ensemble prompting and reverse verification.
\newblock \emph{arXiv preprint arXiv:2502.11520}, 2025.

\bibitem[Team(2025)]{seed2025seed-oss}
ByteDance~Seed Team.
\newblock Seed-oss open-source models.
\newblock \url{https://github.com/ByteDance-Seed/seed-oss}, 2025.

\bibitem[Team et~al.(2025)Team, Bai, Bao, Chen, Chen, Chen, Chen, Chen, Chen, Chen, et~al.]{team2025kimi}
Kimi Team, Yifan Bai, Yiping Bao, Guanduo Chen, Jiahao Chen, Ningxin Chen, Ruijue Chen, Yanru Chen, Yuankun Chen, Yutian Chen, et~al.
\newblock Kimi k2: Open agentic intelligence.
\newblock \emph{arXiv preprint arXiv:2507.20534}, 2025.

\bibitem[Team(2024)]{qwen2.5}
Qwen Team.
\newblock Qwen2.5: A party of foundation models, September 2024.
\newblock URL \url{https://qwenlm.github.io/blog/qwen2.5/}.

\bibitem[Wang et~al.(2023)Wang, Li, Shao, Xu, Dai, Li, Chen, Wu, and Sui]{wang2023math}
Peiyi Wang, Lei Li, Zhihong Shao, RX~Xu, Damai Dai, Yifei Li, Deli Chen, Yu~Wu, and Zhifang Sui.
\newblock Math-shepherd: Verify and reinforce llms step-by-step without human annotations.
\newblock \emph{arXiv preprint arXiv:2312.08935}, 2023.

\bibitem[Xia et~al.(2025)Xia, Li, Liu, Wu, and Liu]{xia2025evaluating}
Shijie Xia, Xuefeng Li, Yixin Liu, Tongshuang Wu, and Pengfei Liu.
\newblock Evaluating mathematical reasoning beyond accuracy.
\newblock In \emph{Proceedings of the AAAI Conference on Artificial Intelligence}, volume~39, pp.\  27723--27730, 2025.

\bibitem[Yang et~al.(2025)Yang, Li, Yang, Zhang, Hui, Zheng, Yu, Gao, Huang, Lv, et~al.]{yang2025qwen3}
An~Yang, Anfeng Li, Baosong Yang, Beichen Zhang, Binyuan Hui, Bo~Zheng, Bowen Yu, Chang Gao, Chengen Huang, Chenxu Lv, et~al.
\newblock Qwen3 technical report.
\newblock \emph{arXiv preprint arXiv:2505.09388}, 2025.

\bibitem[Yu et~al.(2025)Yu, Li, and Wang]{scaling_flaws_of_verifier_guided_yu2025scaling}
Fei Yu, Yingru Li, and Benyou Wang.
\newblock Scaling flaws of verifier-guided search in mathematical reasoning.
\newblock \emph{arXiv preprint arXiv:2502.00271}, 2025.

\bibitem[Zeng et~al.(2025)Zeng, Lv, Zheng, Hou, Chen, Xie, Wang, Yin, Zeng, Zhang, et~al.]{zeng2025glm}
Aohan Zeng, Xin Lv, Qinkai Zheng, Zhenyu Hou, Bin Chen, Chengxing Xie, Cunxiang Wang, Da~Yin, Hao Zeng, Jiajie Zhang, et~al.
\newblock Glm-4.5: Agentic, reasoning, and coding (arc) foundation models.
\newblock \emph{arXiv preprint arXiv:2508.06471}, 2025.

\bibitem[Zeng et~al.(2023)Zeng, Chen, Liu, Jiang, and Jia]{zeng2023mr}
Zhongshen Zeng, Pengguang Chen, Shu Liu, Haiyun Jiang, and Jiaya Jia.
\newblock Mr-gsm8k: A meta-reasoning benchmark for large language model evaluation.
\newblock \emph{arXiv preprint arXiv:2312.17080}, 2023.

\bibitem[Zeng et~al.(2024)Zeng, Liu, Wan, Li, Chen, Dai, Yao, Xu, Qi, Zhao, et~al.]{zeng2024mr}
Zhongshen Zeng, Yinhong Liu, Yingjia Wan, Jingyao Li, Pengguang Chen, Jianbo Dai, Yuxuan Yao, Rongwu Xu, Zehan Qi, Wanru Zhao, et~al.
\newblock Mr-ben: A comprehensive meta-reasoning benchmark for large language models.
\newblock \emph{arXiv e-prints}, pp.\  arXiv--2406, 2024.

\bibitem[Zha et~al.(2025)Zha, Gao, Shen, Hong, Boning, and Katabi]{zha2025rl}
Kaiwen Zha, Zhengqi Gao, Maohao Shen, Zhang-Wei Hong, Duane~S Boning, and Dina Katabi.
\newblock Rl tango: Reinforcing generator and verifier together for language reasoning.
\newblock \emph{arXiv preprint arXiv:2505.15034}, 2025.

\bibitem[Zhang et~al.(2025{\natexlab{a}})Zhang, Hosseini, Bansal, Kazemi, Kumar, and Agarwal]{zhang2025generative}
Lunjun Zhang, Arian Hosseini, Hritik Bansal, Mehran Kazemi, Aviral Kumar, and Rishabh Agarwal.
\newblock Generative verifiers: Reward modeling as next-token prediction.
\newblock In \emph{The Thirteenth International Conference on Learning Representations}, 2025{\natexlab{a}}.
\newblock URL \url{https://openreview.net/forum?id=Ccwp4tFEtE}.

\bibitem[Zhang et~al.(2023)Zhang, Li, Zong, Ying, He, and Qiu]{zhang2023evaluating}
Xiaotian Zhang, Chunyang Li, Yi~Zong, Zhengyu Ying, Liang He, and Xipeng Qiu.
\newblock Evaluating the performance of large language models on gaokao benchmark.
\newblock \emph{arXiv preprint arXiv:2305.12474}, 2023.

\bibitem[Zhang et~al.(2025{\natexlab{b}})Zhang, Zheng, Wu, Zhang, Lin, Yu, Liu, Zhou, and Lin]{zhang2025lessons}
Zhenru Zhang, Chujie Zheng, Yangzhen Wu, Beichen Zhang, Runji Lin, Bowen Yu, Dayiheng Liu, Jingren Zhou, and Junyang Lin.
\newblock The lessons of developing process reward models in mathematical reasoning.
\newblock \emph{arXiv preprint arXiv:2501.07301}, 2025{\natexlab{b}}.

\bibitem[Zheng et~al.(2024{\natexlab{a}})Zheng, Zhang, Zhang, Lin, Lu, Yu, Liu, Zhou, and Lin]{zheng2024processbench}
Chujie Zheng, Zhenru Zhang, Beichen Zhang, Runji Lin, Keming Lu, Bowen Yu, Dayiheng Liu, Jingren Zhou, and Junyang Lin.
\newblock Processbench: Identifying process errors in mathematical reasoning.
\newblock \emph{arXiv preprint arXiv:2412.06559}, 2024{\natexlab{a}}.

\bibitem[Zheng et~al.(2023)Zheng, Chiang, Sheng, Zhuang, Wu, Zhuang, Lin, Li, Li, Xing, et~al.]{mtbench_zheng2023judging}
Lianmin Zheng, Wei-Lin Chiang, Ying Sheng, Siyuan Zhuang, Zhanghao Wu, Yonghao Zhuang, Zi~Lin, Zhuohan Li, Dacheng Li, Eric Xing, et~al.
\newblock Judging llm-as-a-judge with mt-bench and chatbot arena.
\newblock \emph{Advances in neural information processing systems}, 36:\penalty0 46595--46623, 2023.

\bibitem[Zheng et~al.(2024{\natexlab{b}})Zheng, Yin, Xie, Sun, Huang, Yu, Cao, Kozyrakis, Stoica, Gonzalez, et~al.]{zheng2024sglang}
Lianmin Zheng, Liangsheng Yin, Zhiqiang Xie, Chuyue~Livia Sun, Jeff Huang, Cody~Hao Yu, Shiyi Cao, Christos Kozyrakis, Ion Stoica, Joseph~E Gonzalez, et~al.
\newblock Sglang: Efficient execution of structured language model programs.
\newblock \emph{Advances in neural information processing systems}, 37:\penalty0 62557--62583, 2024{\natexlab{b}}.

\bibitem[Zhou et~al.(2025{\natexlab{a}})Zhou, Xu, Zhou, Singh, Gui, and Joty]{zhou2025variation}
Yefan Zhou, Austin Xu, Yilun Zhou, Janvijay Singh, Jiang Gui, and Shafiq Joty.
\newblock Variation in verification: Understanding verification dynamics in large language models.
\newblock \emph{arXiv preprint arXiv:2509.17995}, 2025{\natexlab{a}}.

\bibitem[Zhou et~al.(2025{\natexlab{b}})Zhou, Xu, Wang, Xiong, and Joty]{zhou2025evaluating}
Yilun Zhou, Austin Xu, Peifeng Wang, Caiming Xiong, and Shafiq Joty.
\newblock Evaluating judges as evaluators: The jetts benchmark of llm-as-judges as test-time scaling evaluators.
\newblock \emph{arXiv preprint arXiv:2504.15253}, 2025{\natexlab{b}}.

\bibitem[Zou et~al.(2025)Zou, Yang, Gu, Qiu, Shen, He, and Wang]{zou2025reasonflux}
Jiaru Zou, Ling Yang, Jingwen Gu, Jiahao Qiu, Ke~Shen, Jingrui He, and Mengdi Wang.
\newblock Reasonflux-prm: Trajectory-aware prms for long chain-of-thought reasoning in llms.
\newblock \emph{arXiv preprint arXiv:2506.18896}, 2025.

\end{thebibliography}
\bibliographystyle{iclr2026_conference}

\appendix
\newpage
\section*{Appendix}



\section{Detailed Dataset Sources}

In table ~\Cref{Tab:olympiad_stats} we provide the distribution of the 80 problems we sourced from different Olympiads along with the date the Olympiads were conducted. For the IMO-shortlist, we report the earliest date that the shortlist questions were made publicly available, typically the calendar year \textit{after} the Olympiad was conducted.
\begin{table}[h!]
\centering
\begin{tabular}{@{}llc@{}}
\toprule
\textbf{Contest} & \textbf{Date of Olympiad} & \textbf{\# Questions} \\ \midrule
\href{https://www.imo-official.org/problems/IMO2023SL.pdf}{IMO - Shortlist 2023}  & 21 July 2024 & 10 \\
\href{https://www.imo-official.org/problems/IMO2024SL.pdf}{IMO - Shortlist 2024}  & 23 July 2025 & 29 \\
\href{https://maa.org/putnam/}{Putnam} & 7 Dec 2024 & 12 \\
\href{https://www.egmo.org/}{EGMO (European Girls' Mathematical Olympiad)} & 17 April 2025 & 6 \\
\href{https://www.imo-official.org/year_info.aspx?year=2025}{IMO (International Mathematical Olympiad)}  & 20 July 2025 & 6 \\
\href{https://bmos.ukmt.org.uk/}{BMO (British Mathematical Olympiad)} & 22 Jan 2025 & 4 \\
\href{https://cms.math.ca/competitions/cmo/}{CMO (Canadian Mathematical Olympiad)} & 6 March 2025 & 4 \\
\href{https://maa.org/student-programs/youth-program-awards-winners}{USA-JMO (Junior Mathematical Olympiad)} & 20 March 2025 & 4 \\
\href{https://olympiads.hbcse.tifr.res.in/mathematical-olympiad-2024-2025/}{INMO (Indian National Mathematical Olympiad)} & 19 Jan 2025 & 3 \\
\href{https://maa.org/student-programs/youth-program-awards-winners}{USAMO (United States of America Mathematical Olympiad)} & 20 March 2025 & 2 \\ \midrule
\textbf{Total} &  & \textbf{80} \\ \bottomrule
\end{tabular}
\label{Tab:olympiad_stats}
\caption{Distribution of questions from various Olympiads with Year-wise Splits}
\end{table}

\begin{figure}[b!]
    \centering
    \includegraphics[width=0.8\linewidth]{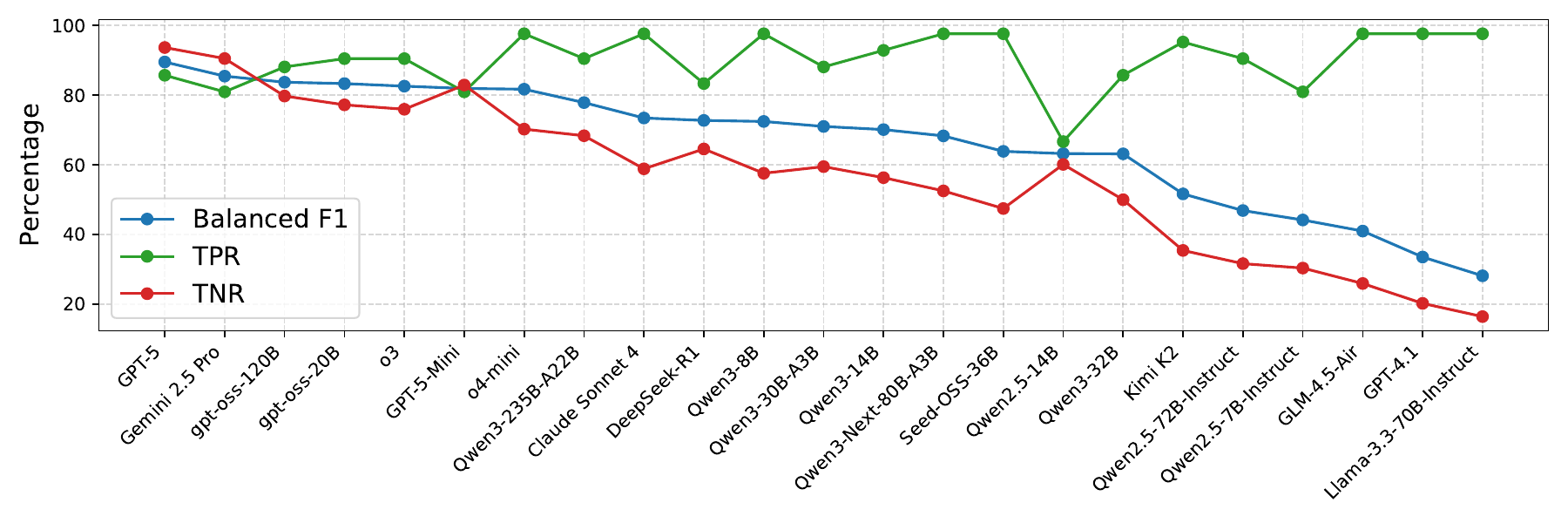}
    \includegraphics[width=0.8\linewidth]{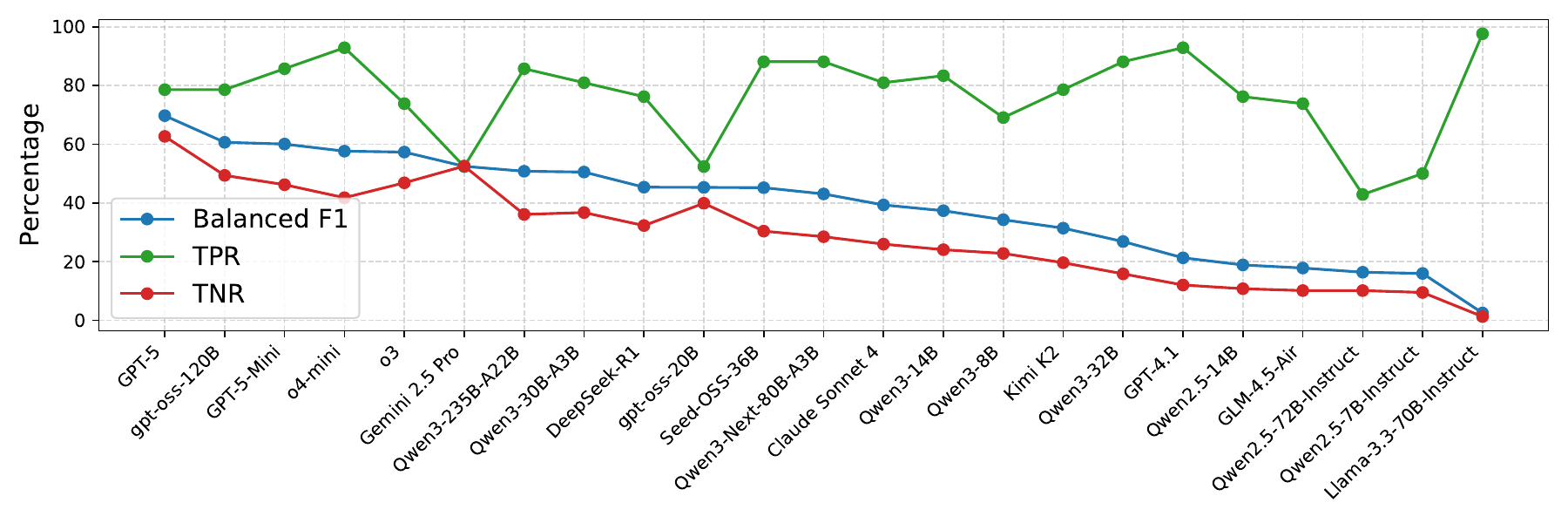}
    \caption{\resplevel and \errorid tasks follow similar trends in TPR and TNR, with weaker verifiers unable to identify mistakes.}
    \label{fig:tpr_tnr_other_tasks}
\end{figure}

\section{Additional Experimental Results}\label{app:additional_results}
We report TPR and TNR for all evaluated models in~\Cref{tab:full_results}, alongside our aggregate metrics presented in~\Cref{sec:exps}. We also visualize TPR and TNR trends for the \resplevel and \steplevel tasks, similar to~\Cref{fig:tpr_tnr_plot}. As shown in~\Cref{fig:tpr_tnr_other_tasks}, TNR is the primary driver in poor Balanced F1 performance: The weaker the verifier, the more it struggles in identifying mistakes, opting to mark nearly every step as correct.

\begin{table}[t!]
\caption{Complete metrics for our three evaluation tasks, reporting Balanced Accuracy, Balanced F1, TPR, and TNR.}
\label{tab:full_results}
\resizebox{\textwidth}{!}{%
\begin{tabular}{lcccccccccccc}
 \toprule
& \multicolumn{4}{c}{\steplevel} & \multicolumn{4}{c}{\resplevel} & \multicolumn{4}{c}{\errorid} \\
& TPR & TNR & Bal. Accuracy & Bal. F1 & TPR & TNR & Bal. Accuracy & Bal. F1 & TPR & TNR & Bal. Accuracy & Bal. F1 \\
\midrule
\rowcolor{lightgray!50}\multicolumn{13}{c}{\textit{\textbf{Generative Critics}, proprietary models}} \\
\midrule
GPT-5 & 94.35 & 78.72 & 86.53 & 85.83 & 85.71 & 93.67 & 89.69 & 89.52 & 78.57 & 62.66 & 70.61 & 69.72 \\ 
Gemini 2.5 Pro & 88.15 & 78.59 & 83.37 & 83.09 & 80.95 & 90.51 & 85.73 & 85.46 & 52.38 & 52.53 & 52.46 & 52.46 \\ 
Claude Sonnet 4 & 97.50 & 43.72 & 70.61 & 60.37 & 97.62 & 58.86 & 78.24 & 73.44 & 80.95 & 25.95 & 53.45 & 39.30 \\ 
GPT-5-Mini & 94.81 & 67.31 & 81.06 & 78.73 & 80.95 & 82.91 & 81.93 & 81.92 & 85.71 & 46.20 & 65.96 & 60.04 \\ 
o3 & 95.09 & 62.31 & 78.70 & 75.29 & 90.48 & 75.95 & 83.21 & 82.58 & 73.81 & 46.84 & 60.32 & 57.31 \\ 
o4-Mini & 97.50 & 52.31 & 74.90 & 68.09 & 97.62 & 70.25 & 83.94 & 81.71 & 92.86 & 41.77 & 67.31 & 57.62 \\ 
GPT-4.1 & 98.24 & 14.10 & 56.17 & 24.66 & 97.62 & 20.25 & 58.94 & 33.55 & 92.86 & 12.03 & 52.44 & 21.29 \\ 
\midrule
\rowcolor{lightgray!50}\multicolumn{13}{c}{\textit{\textbf{Generative Critics}, large ($\geq70$B) models}} \\
\midrule
Kimi K2 & 96.02 & 27.56 & 61.79 & 42.83 & 95.24 & 35.44 & 65.34 & 51.66 & 78.57 & 19.62 & 49.10 & 31.40 \\ 
DeepSeek-R1 & 90.28 & 47.56 & 68.92 & 62.30 & 83.33 & 64.56 & 73.95 & 72.75 & 76.19 & 32.28 & 54.23 & 45.35 \\ 
Qwen3-235B-A22B & 97.41 & 47.69 & 72.55 & 64.03 & 90.48 & 68.35 & 79.42 & 77.87 & 85.71 & 36.08 & 60.90 & 50.78 \\ 
Qwen3-Next-80B-A3B & 97.87 & 37.95 & 67.91 & 54.69 & 97.62 & 52.53 & 75.08 & 68.31 & 88.10 & 28.48 & 58.29 & 43.05 \\ 
Qwen2.5-72B-Instruct & 96.76 & 15.26 & 56.01 & 26.36 & 90.48 & 31.65 & 61.06 & 46.89 & 42.86 & 10.13 & 26.49 & 16.38 \\ 
GLM-4.5-Air & 97.50 & 17.31 & 57.40 & 29.40 & 97.62 & 25.95 & 61.78 & 41.00 & 73.81 & 10.13 & 41.97 & 17.81 \\ 
gpt-oss-120B & 94.54 & 61.67 & 78.10 & 74.64 & 88.10 & 79.75 & 83.92 & 83.71 & 78.57 & 49.37 & 63.97 & 60.64 \\ 
Llama-3.3-70B-Instruct & 98.43 & 10.13 & 54.28 & 18.37 & 97.62 & 16.46 & 57.04 & 28.16 & 97.62 & 1.27 & 49.44 & 2.50 \\ 
\midrule
\rowcolor{lightgray!50}\multicolumn{13}{c}{\textit{\textbf{Generative Critics}, small/medium ($<70$B) models}} \\
\midrule
Qwen3-32B & 91.94 & 36.03 & 63.99 & 51.77 & 85.71 & 50.00 & 67.86 & 63.16 & 88.10 & 15.82 & 51.96 & 26.83 \\ 
Qwen3-30B-A3B & 95.65 & 45.77 & 70.71 & 61.91 & 88.10 & 59.49 & 73.79 & 71.02 & 80.95 & 36.71 & 58.83 & 50.51 \\ 
ByteDance Seed-OSS-36B & 97.04 & 36.54 & 66.79 & 53.09 & 97.62 & 47.47 & 72.54 & 63.88 & 88.10 & 30.38 & 59.24 & 45.18 \\ 
gpt-oss-20B & 93.06 & 57.31 & 75.18 & 70.93 & 90.48 & 77.22 & 83.85 & 83.32 & 52.38 & 39.87 & 46.13 & 45.28 \\ 
Qwen3-14B & 94.17 & 36.79 & 65.48 & 52.91 & 92.86 & 56.33 & 74.59 & 70.12 & 83.33 & 24.05 & 53.69 & 37.33 \\ 
Qwen3-8B & 92.96 & 37.56 & 65.26 & 53.51 & 97.62 & 57.59 & 77.61 & 72.45 & 69.05 & 22.78 & 45.92 & 34.26 \\ 
Qwen2.5-14B-Instruct & 88.33 & 32.56 & 60.45 & 47.59 & 66.67 & 60.13 & 63.40 & 63.23 & 76.19 & 10.76 & 43.47 & 18.86 \\ 
Qwen2.5-7B-Instruct & 84.44 & 13.21 & 48.82 & 22.84 & 80.95 & 30.38 & 55.67 & 44.18 & 50.00 & 9.49 & 29.75 & 15.96 \\ 
\midrule
\rowcolor{lightgray!50}\multicolumn{13}{c}{\textit{\textbf{Process Reward Models}, open-source models}} \\
\midrule
Qwen2.5-Math-PRM-72B & 89.50 & 22.14 & 55.82 & 35.50 & 55.56 & 78.05 & 66.80 & 64.91 & 55.56 & 28.05 & 41.80 & 37.28 \\ 
Qwen2.5-Math-PRM-7B & 87.13 & 27.99 & 57.56 & 42.37 & 44.44 & 81.71 & 63.08 & 57.57 & 44.44 & 25.61 & 35.03 & 32.50 \\ 
Skywork-PRM-7B & 51.55 & 25.50 & 38.52 & 34.12 & 17.65 & 95.89 & 56.77 & 29.81 & 17.65 & 5.48 & 11.56 & 8.36 \\ 
Skywork-PRM-1.5B & 74.53 & 7.08 & 40.81 & 12.94 & 11.76 & 93.15 & 52.46 & 20.89 & 11.76 & 5.48 & 8.62 & 7.48 \\ 
ReasonFlux-PRM-7B & 93.47 & 12.72 & 53.09 & 22.40 & 66.67 & 45.12 & 55.89 & 53.82 & 66.67 & 18.29 & 42.48 & 28.71 \\ 
UniversalPRM-7B & 80.00 & 48.35 & 64.17 & 60.27 & 27.78 & 81.71 & 54.74 & 41.46 & 27.78 & 24.39 & 26.08 & 25.97 \\
\bottomrule
\end{tabular}%
}
\end{table}

\section{Prompts for Generation and Evaluation}
In this section we provide prompts used for generating responses to Olympiad-level math questions. We also provide the prompts used for the \steplevel and \errorid tasks.

\begin{rubricbox}{Prompt used to generate responses to Olympiad questions}\label{prompt:generation}
You are a careful, rigorous math proof assistant. Provide correct, detailed, and complete proofs. 

Solve the following math problem formally. Return a detailed and formal solution that can be verified by a grader.

Use start the proof with <start> followed by each step with <step>...</step>, and end with <end>. 

Only return the solution, in detailed steps, no headers, no explanations, no other text, only the <start> <step>...</step> <step>...</step> ... <end> tags.

\end{rubricbox}

\newpage

\begin{rubricbox}{Prompt used for the \steplevel task}\label{prompt:evaluation}
The following is a math problem and a solution (split into steps, enclosed with tags and indexed from 0):

[Math Problem]

\{problem\}

[Solution]

\{steps\}

Your task is to review and critique the solution step-by-step.

 For each step, determine if it is correct or incorrect.
A correct step is one where all of the content is correct, and is logically consistent with all previous steps and information given in the problem.

An incorrect step is one where the content is incorrect, or is not logically consistent with all previous steps and information given in the problem, or is based on an error in a previous step.

Important: Any step that contains or is based on an error is considered incorrect. That is, if the error is carried forward from a previous step or is based on an error in the previous step, consider the step incorrect.

Provide reasoning for your correctness determinations. Your final verdict should be a comma-separated list of yes and no's, where each yes or no corresponds to a step's correctness, with yes meaning correct and no meaning incorrect.

Please use the following format to return your answer:

 Reasoning: <your reasoning for each step>
 
 Verdict: <your comma-separated list of yes and no's>
 
Do not use any other formatting, including markdown, bold text, code blocks, or any other formatting. If your formatting is incorrect, your evaluation will be affected.

\end{rubricbox}

\begin{rubricbox}{Prompt used for the \errorid task}\label{prompt:error_id}
The following is a math problem and a solution (split into steps, enclosed with tags and indexed from 0):

[Math Problem]

\{problem\}

[Solution]

\{steps\}

Your task is to identify the first incorrect step in the solution.

Instructions:

- Review each step carefully for mathematical correctness and logical consistency

- A step is incorrect if it contains mathematical errors, logical inconsistencies, or is based on errors from previous steps

- Find the FIRST step that is incorrect (0-indexed)

- If ALL steps are correct, return -1

Provide your reasoning and then give your final answer as a single number in the specified format.

Please use the following format to return your answer:

Reasoning: <your detailed reasoning explaining which steps are correct/incorrect and why>

Verdict: <the step number of the first incorrect step or -1 if all steps are correct>

Examples:

- If step 0 is the first incorrect step: 0

- If step 3 is the first incorrect step: 3

- If all steps are correct: -1

Do not use any other formatting, including markdown, bold text, code blocks, or any other formatting. If your formatting is incorrect, your evaluation will be affected.
    
\end{rubricbox}

\section{Annotation details}\label{app:annotator}
Each sample was annotated over four rounds: An initial annotation round and three rounds of reviews to resolve disagreements. A total of 52 annotators were employed for grading, with 35 having at least a graduate degree in mathematics or related fields. On average, a model response took 90 minutes to grade and 63 minutes to review, with the longest response taking up to 4 hours. Annotators were given access to external tools, such as the internet, python, Wolfram Mathematica, and LLMs strictly as assistive aids. 

We present the detailed annotation guidelines provided to the math experts for step-by-step evaluation of each model solution below.

\begin{rubricbox}{Annotation instructions to human annotators}\label{prompt:annotation_instructions}

When annotating, refer to the reference answer(s) as possible solution(s)/proof(s). Each question may have multiple valid approaches, as these are open-ended questions. The provided reference answer(s) is an example of a valid approach; it may not be the only such valid approach.

Base your correctness decision off of the following criteria:

Correct: A step is considered correct if it is: 

Computationally valid: There are no mistakes in rote mathematical operations, such as addition or computing values of known functions (e.g., sin(pi/2))

Logically valid: The step follows logically from previous steps and information present in the original question. There are no intermediate mistakes in the reasoning. Any and all conclusions in the step must be logically deducible from previous correct steps.

If a step invokes any third-party mathematical results, such as known theorems / lemmas (e.g., fundamental theorem of calculus) or intermediate results from previous steps, then annotators must verify that the result is used in a valid way:

(1) all assumptions of the result (theorem) are met

(2) the consequence of the result (theorem) is correctly described and applied to the specific problem

Important: Do not apply “Error carried forward” grading. 

If a current step is derived from a previous step that is incorrect, consider the current step incorrect, even if the logic/computation of the step is correct. 

Example: 

Step 1: 1 + 1 = 3 [Incorrect]

Step 2: We now must add 5 to Step 1’s result, which gives us 8 [Incorrect, even though the computation in the step is correct; It is based on an incorrect Step 1]

Extra note:

“Hand-waviness”: If a model produces a “hand-wavy” argument, wherein they say that a new result follows by similar logic/computation as a previously established result, then annotators must verify that the hand-wavy argument in-fact holds. This means verifying
(1) The previously established result’s assumptions are met by the new result scenario
(2) The previously established computation/logic is applicable to the new

Example: 

Step N: A valid proof of Case 1, yielding Result 1

Step N+1: Case 2 follows by a similar argument to Case 1, yielding Result 2. 

[This is “hand-wavy”, as the exact computation is omitted by appealing to previously computed Steps]

Incorrect: A step is considered incorrect if it is:

Based in any way on an incorrect past step.

Logically invalid: The model’s output contains a reasoning error or mistake. Examples:
Unfounded logical leap 

Incorrectly invoking a mathematical result or past result when assumptions/conditions are not satisfied

Incorrect application of a mathematical result when conditions are met, i.e., mis-applying a theorem.

Failing to consider/cover a scenario or case within a proof, i.e., the proof concludes without covering all scenarios and is incomplete.

If the top-level proof misses a case/scenario: As this case involves text not in the model output, there is no concrete step to mark as incorrect. As a result, mark the conclusion of the proof (i.e., last step) as incorrect and provide corresponding justification.

If an intermediate result is stated, but the derivation of the intermediate result misses a case/scenario: Mark the step that states the intermediate result as incorrect (as well as any subsequent steps that depend on the intermediate result).  As a concrete toy example
Say a model is doing Proof by Cases for all real numbers. 

It splits its analysis into 2 cases, Case 1 (positives) and Case 2 (negatives). 
For Case 1, it proves the claim for all positive integers, but does not consider non-integer reals. 

Mark the step that contains the conclusion of Case 1 incorrect, as well as any subsequent steps that depend on Case 1.

Computationally invalid: Makes an operation / value computation mistake. This should be relatively easy to spot, but please verify all complex expressions, such as integrals, trigonometric functions, etc. 

Note: This is not an exhaustive list of errors. Verify all computations, and document any error that occurs, no matter how minor.

\end{rubricbox}

\section{Evaluated baselines}\label{app:baselines}
Here we provide a comprehensive list of models that were evaluated on our benchmark.
\begin{itemize}
    \item OpenAI: GPT-5, GPT-5-Mini~\citep{gpt5}, o3, o4-Mini~\citep{o4mini}, GPT-4.1~\citep{openai_gpt41_2025}, gpt-oss-120b, gpt-oss-20b~\citep{gptoss_agarwal2025}
    \item Google: Gemini 2.5 Pro~\citep{gemini_25}
    \item Anthropic: Claude Sonnet 4~\citep{claude4}
    \item Moonshot (Kimi): Kimi-K2-Instruct-0905~\citep{team2025kimi} \item DeepSeek: DeepSeek-R1~\citep{guo2025deepseek}
    \item Alibaba Qwen: Qwen3-235-A22B, Qwen3-Next-80B-A3B, Qwen3-32B, Qwen3-30B-A3B, Qwen3-14B, Qwen3-8B~\citep{yang2025qwen3}, Qwen2.5-72B-Instruct, Qwen2.5-14B-Instruct, Qwen2.5-7B-Instruct~\citep{qwen2.5}, Qwen2.5-Math-PRM-72B, Qwen2.5-Math-PRM-7B~\citep{zhang2025lessons} 
    \item Zhipu GLM: GLM-4.5-Air~\citep{zeng2025glm}
    \item Meta: Llama-3.3-70B-Instruct~\citep{grattafiori2024llama}
    \item ByteDance: ByteDance Seed-OSS-36B~\citep{seed2025seed-oss}
    \item Skywork: Skywork-PRM-7B, Skywork-PRM-1.5B~\citep{he_2024_16998085} 
    \item ReasonFlux-PRM-7B~\citep{zou2025reasonflux}
    \item UniversalPRM-7B~\citep{tan2025aurora}
\end{itemize}

For Kimi K2, DeepSeek-R1, and GLM-4.5-Air, we used \url{together.ai} for inference. All other open-weight baselines were run locally, hosted via vLLM~\citep{kwon2023efficient} or SGLang~\citep{zheng2024sglang}.

\subsection{PRM Threshold Tuning}\label{app:prm:threshold}
To decide the cutoff threshold for evaluated PRMs, we select 100 responses at random from our benchmark and tune PRM performance against this subset, following~\citep{zheng2024processbench}. The same 100 responses are kept fixed across all baselines, and we sweep the threshold from 0.1 to 0.9 in increments of 0.05. To select the threshold, we compute the harmonic mean of the three task-specific Balanced F1 Scores, prioritizing selecting a threshold that yields strong yet balanced performance. We find that PRM performance can vary considerably based on chosen threshold.

\end{document}